\documentclass[10pt]{article} 
\usepackage[preprint]{tmlr}

\usepackage{amsmath,amsfonts,bm}

\def\eqref#1{equation~\ref{#1}}

\def\1{\bm{1}}

\DeclareMathAlphabet{\mathsfit}{\encodingdefault}{\sfdefault}{m}{sl}
\SetMathAlphabet{\mathsfit}{bold}{\encodingdefault}{\sfdefault}{bx}{n}

\usepackage{hyperref}
\usepackage{url}

\usepackage{amsmath,amsthm, amssymb, amsfonts}
\usepackage{graphicx}  
\usepackage{wrapfig}  
\usepackage{soul}
\usepackage{multirow}
\usepackage{graphicx}
\usepackage{caption}
\usepackage{subcaption}
\usepackage{tabularx}  
\usepackage{booktabs}
\usepackage{svg}
\usepackage{thmtools, thm-restate}
\usepackage[figurewithin = section]{caption}
\usepackage{booktabs}
\usepackage{array}
\usepackage{floatrow}
\usepackage{xspace}

\theoremstyle{definition}
\newtheorem{definition}{Definition}[section]

\newcommand{\quotes}[1]{``#1''}

\newcommand{\modelname}{\textsc{ARK}\xspace}
\newcommand{\modellong}{Autoregressive Knowledge generation\xspace}

\newcommand{\modelvae}{\textsc{SAIL}\xspace}

\newcommand{\transformerark}{$t$-\textsc{ARK}\xspace}

\newcommand{\transformerbaselineabbr}{$t$-\textsc{SAIL}\xspace}

\newcommand{\repolink}{\url{https://github.com/thiviyanT/ARK}}  

\makeatletter
\def\thickhline{%
  \noalign{\ifnum0=`}\fi\hrule \@height \thickarrayrulewidth \futurelet
   \reserved@a\@xthickhline}
\def\@xthickhline{\ifx\reserved@a\thickhline
               \vskip\doublerulesep
               \vskip-\thickarrayrulewidth
             \fi
      \ifnum0=`{\fi}}
\makeatother

\newlength{\thickarrayrulewidth}
\title{
    Autoregressive Models for Knowledge Graph Generation
}

\author{
\name Thiviyan Thanapalasingam\thanks{Equal contribution.} \email thiviyan.t@gmail.com \\
\addr Universiteit van Amsterdam \\
1098XH Amsterdam, The Netherlands
\AND
\name Antonis Vozikis\footnotemark[1] \email a.vozikis@uva.nl \\
\addr Universiteit van Amsterdam \\
1098XH Amsterdam, The Netherlands
\AND
\name Peter Bloem \email p.bloem@vu.nl \\
\addr Vrije Universiteit Amsterdam \\
1081HV Amsterdam, The Netherlands
\AND
\name Paul Groth \email p.t.groth@uva.nl \\
\addr Universiteit van Amsterdam \\
1098XH Amsterdam, The Netherlands
}

\usepackage{xcolor} 

\def\month{MM}  
\def\year{YYYY} 
\def\openreview{\url{https://openreview.net/forum?id=XXXX}} 

\begin{document}

\maketitle

\begin{abstract}
Knowledge Graph (KG) generation requires models to learn complex semantic dependencies between triples while maintaining domain validity constraints. Unlike link prediction, which scores triples independently, generative models must capture interdependencies across entire subgraphs to produce semantically coherent structures. We present \textbf{ARK} (\textbf{A}uto-\textbf{R}egressive \textbf{K}nowledge Graph Generation), a family of autoregressive models that generate KGs by treating graphs as sequences of (head, relation, tail) triples. ARK learns implicit semantic constraints directly from data, including type consistency, temporal validity, and relational patterns, without explicit rule supervision. On the IntelliGraphs benchmark, our models achieve 89.2\% to 100.0\% semantic validity across diverse datasets while generating novel graphs not seen during training. We also introduce \textbf{SAIL}, a variational extension of ARK that enables controlled generation through learned latent representations, supporting both unconditional sampling and conditional completion from partial graphs. Our analysis reveals that model capacity (hidden dimensionality $\geq$ 64) is more critical than architectural depth for KG generation, with recurrent architectures achieving comparable validity to transformer-based alternatives while offering substantial computational efficiency. These results demonstrate that autoregressive models provide an effective framework for KG generation, with practical applications in knowledge base completion and query answering. Our code is available on \repolink. 
\end{abstract}

\section{Introduction}
\label{intro}

Knowledge Graphs (KGs) encode knowledge as graphs of entities connected by typed relations, powering applications from search engines to drug discovery \citep{hogan2021knowledge}. However, even large-scale KGs such as Wikidata miss substantial world knowledge. Although Knowledge Graph Embedding (KGE) models address incompleteness \citep{bordes2013translating, yang2014embedding}, they score each triple independently, failing to capture the interdependencies that define valid knowledge structures. This independence assumption becomes particularly problematic for complex facts requiring multiple related triples to represent accurately \citep{nathani-etal-2019-learning}. The KG generation task presents three key challenges that distinguish it from existing KG modeling approaches: (1) \textit{Semantic constraint satisfaction}: generated triples must collectively satisfy domain rules (temporal consistency, type constraints) without explicit supervision, (2) \textit{Structural coherence}: entities must form connected subgraphs with valid relational patterns, and (3) \textit{Joint distribution modeling}: unlike KGE models, which score each triple independently, we model $p(G)$ over complete graphs.

Consider representing \quotes{\textit{Barack Obama was the US President from 2009-2017}} in a KG; this requires multiple interdependent triples that must also satisfy temporal constraints (start year $\leq$ end year), and type consistencies (only persons can be presidents). Traditional link predictors cannot ensure that these constraints are satisfied collectively, leading to semantically invalid predictions that undermine downstream reasoning tasks \citep{thanapalasingam2023intelligraphs}. This is particularly critical for $N$-ary relations that inherently cannot decompose into independent binary predictions \citep{wen2016representation}. In contrast to link prediction, KG generation addresses these limitations by modeling joint distributions over sets of triples, enabling the sampling of complete graphs that satisfy semantic constraints across all their components simultaneously. 

Generative models can learn these interdependencies by modeling entire (sub)graphs rather than individual links \citep{xie2022genkgc}. Previous work on generative models in the KG domain has primarily focused on generating triples from text \citep{saxena2022sequence,chen-etal-2020-kgpt} or learning embeddings for downstream tasks \citep{xiao-etal-2016-transg, he2015learning}, rather than learning distributions over complete graph structures. To our knowledge, no prior work has demonstrated the ability to sample entire, semantically valid KGs from learned probabilistic models. This raises a fundamental question: \textit{What is required to effectively model $p_{\theta}(G)$ for Knowledge Graphs?} We observe that KGs can be naturally represented as sequences of triples (head, relation, tail), suggesting that autoregressive sequence models may be well-suited for this task.

We introduce \textbf{\modelname} (\textbf{A}uto-\textbf{R}egressive \textbf{K}nowledge Graph Generation), a family of autoregressive models that generate KGs by sequentially predicting triples. Our models learn semantic constraints, including type consistency, temporal validity, and relational patterns, directly from data without explicit rule supervision. On the IntelliGraphs benchmark, ARK achieves 89.2\% to 100.0\% semantic validity across diverse datasets. We further present \textbf{\modelvae} (\textbf{S}equential \textbf{A}uto-Regress\textbf{I}ve Knowledge Graph Generation with \textbf{L}atents), a light-weight probabilistic extension of \modelname that enables controllable generation from learned latent distributions.

We focus on subgraph generation, producing graphs of 3 to 212 triples that satisfy domain constraints. This scope aligns with practical applications, including knowledge base completion, query answering, and data augmentation, where generating valid local structures is the primary requirement. Our contributions are as follows:

\begin{enumerate}
    \item We introduce \modelname, an autoregressive approach to Knowledge Graph generation that learns implicit semantic constraints from data, achieving 89.2\% to 100.0\% validity on the IntelliGraphs benchmark without explicit rule supervision;
    \item We present \modelvae, a variational extension that enables controlled generation through learned latent representations, supporting both unconditional sampling and conditional completion from partial graphs;
    \item We show that model capacity (hidden dimensionality $\geq$ 64) matters more than architectural depth, with single-layer GRUs matching deeper transformer performance while offering computational efficiency;
    \item We release our models and code, establishing baselines for future work on KG generation. Our code is available at \repolink. 
\end{enumerate}

\section{Preliminaries}
\label{sec:preliminaries}

\textbf{Knowledge Graph Generation} \quad We consider the task of generating semantically valid Knowledge Graphs $G = (E,R,T)$ where $E$ is a set of entities, $R$ is a set of relations, and $T \subseteq E \times R \times E$ is a set of triples. Unlike link prediction, which focuses on individual triple classification, our goal is to generate a collection of triples (\emph{i.e.} subgraphs) that satisfy domain-specific semantic constraints while capturing interdependencies. This subgraph inference task is particularly crucial for $N$-ary relations and more complex facts that cannot decompose into independent binary predictions \citep{thanapalasingam2023intelligraphs}. For example, temporal constraints require that \emph{start year} precede \emph{end year} across multiple triples, while type constraints ensure that only valid entity-relation combinations appear together. Models must generate and validate entire structures collectively rather than scoring individual triples. This is distinct from link prediction or generation of triples from text, as the models need to assign probability to and sample entire sets of interdependent triples. We emphasize that we address subgraph generation (coherent collections of 3-212 triples), which is the practical use case for many real-world applications including KG completion and query answering.

\begin{definition}[Knowledge Graph Generation]
\label{def:kg-generation}
Given a training set of Knowledge Graphs $\mathcal{D} = \{G_1, ..., G_n\}$, learn a generative model $p_\theta(G)$ that can sample new graphs $G' \sim p_\theta$ such that $G'$ satisfies semantic validity constraints $\mathcal{S}$ while not appearing in $\mathcal{D}$.
The task is neither link prediction (predicting individual triples) nor text-to-KG extraction (generating KGs from text), but learning to generate complete graph structures that satisfy implicit semantic constraints. We use this for: (1) unconditional sampling (generating diverse valid graphs), (2) conditional completion (completing partial graphs), and (3) learning the underlying distribution for compression.
\end{definition}

\begin{definition}[Semantic Validity]
\label{def:semantic-validity}
A graph $G$ is semantically valid if it satisfies constraints $\mathcal{S} = \{s_1, ..., s_k\}$ where each $s_i$ is a rule (e.g., type constraints, temporal consistency, relational dependencies).
Concrete examples include: (1) start\_year $\leq$ end\_year (temporal consistency), (2) entity types must match relation requirements (e.g., only Person entities can have "birthplace" relations), (3) graphs must form valid paths (connected, acyclic, directional), (4) type consistency (directors are people, genres are categories), (5) similar type constraints plus connectivity. These constraints are not enforced during generation; the model must learn them from data.
\end{definition}

\textbf{Variational Inference} \quad To learn latent representations, we use the $\beta$-VAE framework \citep{kingma2013auto,higgins2017beta}, which aims to maximize the evidence lower bound (ELBO):

\begin{equation}
\label{eq:elbo}
\mathcal{L}(\phi, \theta; G) = \mathbb{E}_{q_\phi(z|G)}[\log p_\theta(G|z)] - \beta\;\text{KL}[q_\phi(z|G) || p(z)] \text{.}
\end{equation}

\section{Sequential Decoding for Knowledge Graph Generation}
\label{sec:sequential-generation}

\begin{figure}[!t]
    \centering
    \includegraphics[width=14cm]{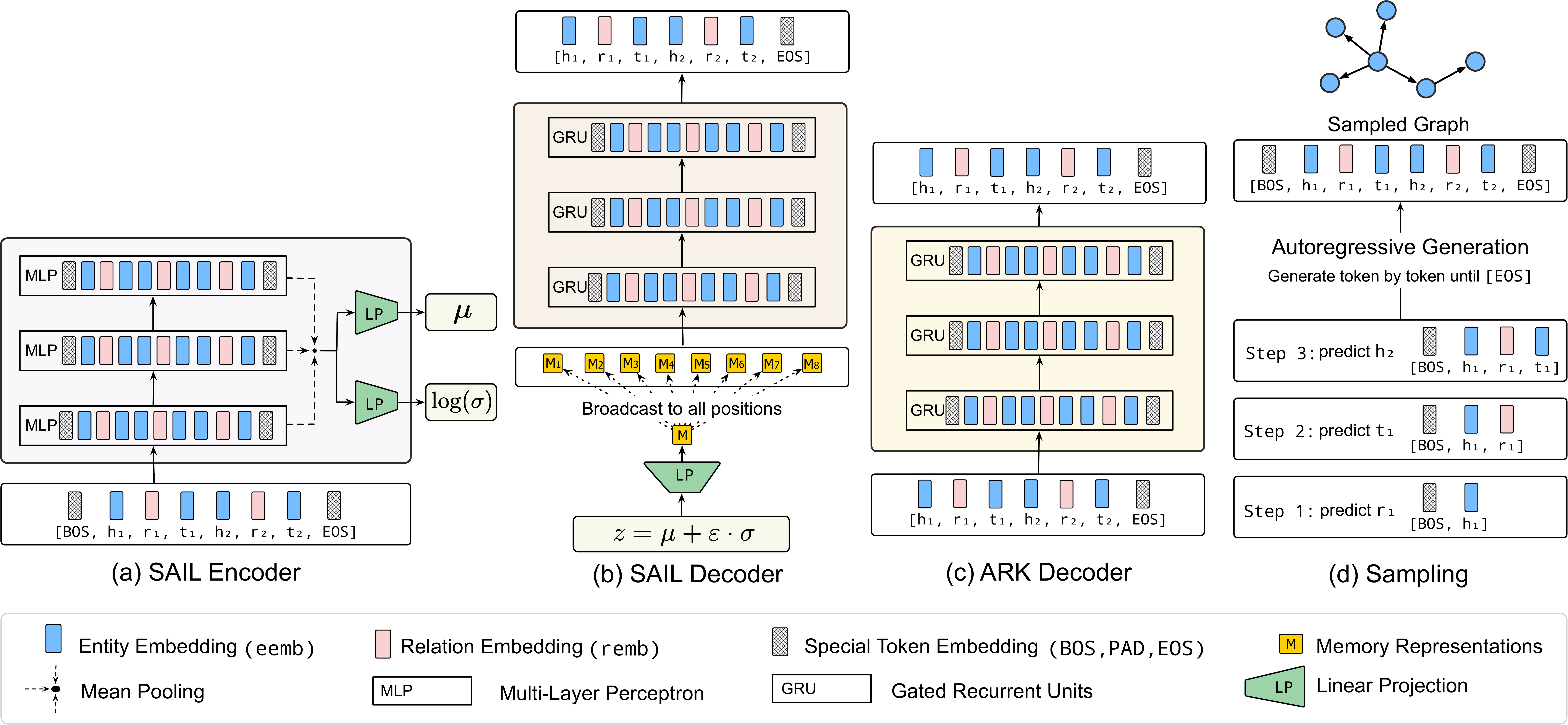}
    \caption{
        Overview of Model Architectures.
        \textbf{(a) \modelvae Encoder:} Multi-layer perceptron (MLP) processes linearized KG sequences $[\texttt{BOS}, h_1, r_1, t_1, h_2, r_2, t_2, \ldots, \texttt{EOS}]$, with mean pooling to produce fixed-size representations. Linear projections generate latent distribution parameters $\mu$ and $\log\sigma$.
        \textbf{(b) \modelvae Decoder:} GRU-based decoder conditions on sampled latent code $z \sim \mathcal{N}(\mu, \sigma^2)$ by broadcasting $z$ to all sequence positions and concatenating with embeddings $[M_1, M_2, \ldots, M_n]$ at each timestep.
        \textbf{(c) \modelname Decoder:} GRU decoder for \modelname operates without latent conditioning, processing embedded sequences directly through stacked GRU layers.
        \textbf{(d) Sampling:} Autoregressive generation proceeds token-by-token with causal masking until \texttt{EOS} token or maximum length.
    }
    \label{fig:model-architecture}
\end{figure}

We present our approach to Knowledge Graph generation through sequential decoding. We first describe how graphs are linearized into token sequences, then introduce \modelname\xspace (\textbf{A}uto-\textbf{R}egressive \textbf{K}nowledge Graph Generation), our autoregressive decoder model, followed by \modelvae (\textbf{S}equential \textbf{A}uto-Regress\textbf{I}ve Knowledge Graph Generation with \textbf{L}atents), which extends \modelname with a variational framework for controlled generation.

\subsection{Graph Input Processing}
To enable sequential generation, we linearize KGs into token sequences. A graph $G$ containing triples $(h_1, r_1, t_1), ..., (h_n, r_n, t_n)$ is represented as $[\texttt{BOS}, h_1, r_1, t_1, h_2, r_2, t_2, ..., h_n, r_n, t_n, \texttt{EOS}]$, where \texttt{BOS} marks the sequence start and \texttt{EOS} indicates termination. These tokens provide explicit generation boundaries, enabling the model to learn proper initiation and termination conditions. We employ a unified vocabulary $\mathcal{V} = \{\texttt{BOS}, \texttt{PAD}, \texttt{EOS}\} \cup \mathcal{E} \cup \mathcal{R}$ that combines special tokens, entities, and relations into a single embedding space. Variable-length graphs are padded to a fixed maximum length $L_{\text{max}}$ using \texttt{PAD} tokens for batched training. During training, triple ordering within sequences is randomized to prevent the model from learning spurious positional patterns.\footnotemark This randomization addresses permutation sensitivity: the model sees each graph in multiple orderings, learning order-invariant semantic constraints rather than positional patterns. Notably, our generated graphs are evaluated order-independently (checking whether the set of triples satisfies constraints), so generation quality does not depend on decoding order.

\footnotetext{This prevents leakage; \emph{e.g.}, in the syn-paths dataset, which models KG paths, linearizing in path order makes the pattern trivially easy to learn.}

\subsection{\modellong (\modelname)}

\modelname is an autoregressive model that generates KGs token-by-token, predicting each element conditioned on all previous tokens. We use Gated Recurrent Units (GRUs) \citep{cho2014properties} as the sequence model, exploiting the natural sequential structure of linearized graphs. See Appendix~\ref{subsec:gru-details} for architectural details.

The model is trained autoregressively with cross-entropy loss, conditioning on ground-truth previous tokens:
$
\mathcal{L}_{\text{\modelname}} = -\sum_{t=1}^{T} \log p(x_t | x_{<t})
$.

\textbf{Generation} \quad During inference, \modelname generates graphs sequentially starting from the $\texttt{BOS}$ token. At each timestep $t$, the model computes the probability distribution $p(x_{t+1} | x_{\leq t})$ over the vocabulary. We select the next token through sampling controlled by temperature and top-$k$. Concretely, we divide logits by temperature $T$, keep only the top-k tokens, then retain the smallest prefix whose cumulative probability mass exceeds $p$ (top-$p$), renormalize and sample one token. Decoding stops on \texttt{EOS} or when the maximum graph length has been reached. The generated sequence is parsed into triples by extracting consecutive $(h, r, t)$ token triplets between $\texttt{BOS}$ and $\texttt{EOS}$ markers, with incomplete triples discarded during post-processing.

\subsection{Sequential Autoregressive Knowledge Graph Generation with Latents (\modelvae)}

\modelvae extends \modelname by incorporating a variational autoencoder framework, similar to \citet{bowman2016generating}, enabling probabilistic generation from learned latent distributions, $z$. This extension allows for controlled generation and interpolation in latent space while maintaining the efficiency of GRU-based decoding. The VAE framework enables: (1) controlled generation through latent manipulation, (2) interpolation between graphs, and (3) conditional generation from partial graphs. The technical challenge is learning meaningful, continuous representations of discrete graph structures.

\textbf{Encoder} \quad The encoder processes the input sequence through a multi-layer perceptron (MLP) to obtain a fixed-size representation.\footnotemark~Each input triple $(h,r,t)$ is embedded as $[E_e[h];E_r[r];E_e[t]] \in \mathbb{R}^{3d}$, and we take a mean over the sequence to form a graph-level vector. The MLP consists of multiple dense layers, with the number of these layers chosen to match the number of stacked GRU layers in the decoder. The encoder then processes the aforementioned sequence through these layers, while ReLU is used as the activation function. The final hidden representation is projected to latent distribution parameters, $\boldsymbol{\mu}$ and $\log \boldsymbol{\sigma}^2$.

\footnotetext{It may seem unusual to use an MLP here, which has no inductive bias for sequential data. In our experiments, a GRU-based encoder performed notably worse. We leave investigation of this counter-intuitive result to future work. Note that the $t$-SAIL architecture \emph{does} provide a sequential inductive bias in both the encoder and decoder.}

\textbf{Latent Sampling} We sample from the latent distribution using the reparameterization trick:
\begin{equation}
\mathbf{z} = \boldsymbol{\mu} + \boldsymbol{\sigma} \odot \boldsymbol{\epsilon}, \quad \boldsymbol{\epsilon} \sim \mathcal{N}(0, \mathbf{I})
\end{equation}
Following standard VAE practice, we use a fixed prior $p(z) = \mathcal{N}(0,I)$ rather than learning it, which acts as a regularizer. The learned components of our model are the encoder $q_\phi(\mathbf{z}|G)$, which maps graphs to latent distributions, and the decoder$p_\theta(G|\mathbf{z})$, which reconstructs graphs from latent codes. Despite using this simple fixed prior, our t-SNE visualizations in Figure 4 demonstrate that the learned posterior captures meaningful structure, with clear clustering by genre.

\textbf{Decoder} \quad The decoder extends \modelname's GRU architecture by conditioning on the latent variables, $\mathbf{z}$. The latent representation is first projected and used to initialize the decoder's hidden state:
$
\mathbf{h}_0 = \tanh(\mathbf{W}_{\text{init}} \mathbf{z} + \mathbf{b}_{\text{init}})
$
To maintain global conditioning throughout generations, $\mathbf{z}$ is broadcast to all sequence positions. At each timestep, we concatenate the projected latent code with the input embedding:
$
\mathbf{x}'_t = [\mathbf{x}_t; \mathbf{W}_z \mathbf{z}]
$
This ensures that the global graph structure encoded in $\mathbf{z}$ influences every token prediction, allowing the decoder to maintain semantic consistency across the entire sequence. \modelvae is trained by maximizing the ELBO (as shown in Equation \ref{eq:elbo}). 

\textbf{Generation \& Sampling} \quad To generate a graph using the model, we sample $\mathbf{z} \sim \mathcal{N}(0, \mathbf{I})$ from the prior distribution. We call this \emph{unconditional generation}. Additionally, we define \emph{conditional generation} where we encode a partial graph to obtain the posterior $q(\mathbf{z}|G_{\text{partial}})$, sample from it, and then complete the sequence.
The generation then follows an autoregressive process where the probability of the complete graph factorizes as:
$
p_\theta(G | \mathbf{z}) = \prod_{t=1}^{T} p_\theta(x_t | x_{<t}, \mathbf{z})
$.
We use beam search with 
$\text{score}(x_{1:t} | \mathbf{z}) = \sum_{i=1}^{t} \log p_\theta(x_i | x_{<i}, \mathbf{z})$. Latent conditioning enables controlled generation by manipulating $\mathbf{z}$, we can interpolate between graphs or explore specific regions of the latent space to generate graphs with desired properties.

\section{Evaluation}
\label{sec:evaluation}

We evaluate a family of RNN and transformer-based models on the IntelliGraphs benchmark \citep{thanapalasingam2023intelligraphs}, which consists of five datasets designed to test different aspects of Knowledge Graph generation. 

\textbf{Benchmark} \quad
IntelliGraphs includes three synthetic datasets (syn-paths, syn-types, syn-tipr) with algorithmically verifiable semantics, ranging from simple path structures to temporal constraints requiring reasoning about time intervals, and two real-world Wikidata-derived datasets (wd-movies, wd-articles) capturing complex relational patterns from movie and academic publication domains. Synthetic datasets contain fixed-size graphs (3-5 triples) with small vocabularies (30-130 entities), while Wikidata datasets feature variable-size graphs (2-212 triples) with large entity vocabularies (24K-61K entities), providing diverse challenges for evaluating generation quality and semantic validity. Detailed dataset characteristics and semantic constraints are provided in Appendix~\ref{app:dataset-details}. To the best of our knowledge, IntelliGraphs is the only benchmark for KG generation, while other KG datasets focus on link prediction. While IntelliGraphs focuses on subgraph generation rather than complete KG generation, this aligns with practical applications where generating coherent subgraphs is the primary requirement, such as knowledge completion and query answering.

\textbf{Baselines} \quad 
The probabilistic baselines from \citet{thanapalasingam2023intelligraphs} decompose graph generation as $p(F) = p(S|E)p(E)$, where $E$ represents entities and $S$ represents structure. The \emph{uniform} baseline samples from uniform distributions, providing estimates for compression bits by assuming equal likelihood for all configurations. The KGE-based baselines (TransE, ComplEx, DistMult) estimate $p(E)$ using entity frequencies with Laplace smoothing and $p(S|E)$ using learned scoring functions: TransE models relations as translations \citep{bordes2013translating}, DistMult uses bilinear interactions \citep{yang2014embedding}, and ComplEx employs complex-valued embeddings \citep{trouillon2016complex}. These models convert scores to probabilities through sigmoid functions and compute compression as $-\log_2 p(S|E) - \log_2 p(E)$. These baselines come directly from  \citet{thanapalasingam2023intelligraphs}, which established them as generation baselines. While the KGE models weren't originally designed for generation, their failure (producing 76-100\% empty graphs) demonstrates precisely why methods designed for independent triple scoring cannot handle joint graph generation. To study architectural choices, we also implement transformer-based variants: \transformerark uses a transformer decoder with causal self-attention, while \transformerbaselineabbr extends this with a variational framework employing transformer encoders and decoders. These variants allow us to examine whether attention mechanisms provide benefits for KG generation. Other approaches like KGT5 \citep{kochsiek-etal-2023-friendly} and recent LLM-based methods address different tasks (text-to-KG extraction, question answering) rather than learning generative distributions over graph structures.

\textbf{Evaluation Metrics} \quad We evaluate generation quality through three primary metrics: (1) \textit{Semantic Validity} -- the proportion of generated graphs that satisfy dataset-specific semantic constraints, measuring whether the model learns to respect domain rules without explicit supervision; (2) \textit{Novelty} -- the proportion of generated graphs not present in the training set, distinguishing genuine generation from memorization; and (3) \textit{Compression} -- the information-theoretic measure $-\log p(G)$ in bits, quantifying how efficiently the model encodes graph structure. For variational models, we additionally report the KL divergence between the approximate posterior and prior. These metrics collectively assess whether models capture the underlying data distribution while maintaining semantic coherence and generalization capability.

\subsection{Compression Code Length}

We express the negative log-likelihood, $-\log_2(p_{\theta})$, in bits-per-graph. See Appendix \ref{subsec:compression_length} for details. This measures both the ability to compress and to predict \citep[Section~3.2]{grunwald2007minimum}.

\textbf{Results} \quad Table \ref{tab:graph-sampling-results} shows the compression performance across all models. \modelname achieves strong compression rates across all datasets, with 27.65 bits for syn-paths (compared to 30.49 bits for the uniform baseline) and 23.48 bits for syn-tipr. On real-world datasets, \modelname achieves the best overall compression with 98.19 bits for wd-movies and 205.24 bits for wd-articles, demonstrating efficient encoding of complex graph structures. While their compression on syn-types is higher (59.63 and 59.79 bits), both models compensate with strong semantic validity in generation tasks. By contrast, the variational models (\modelvae and \transformerbaselineabbr) report ELBO upper bounds rather than exact compression, as they use latent vectors $\mathbf{z}$ to capture graph structure. Their compression includes both reconstruction and KL divergence terms, with the KL component varying from nearly zero to syn-types (0.15 bits) to moderate values on other datasets (13-32 bits), indicating adaptive latent space usage on dataset complexity. 

\subsection{Sampling from Latent Variable, $z$}

\begin{table}[!t]
\caption{Semantic validity and compression length in bits of the graphs generated. We sample graphs and check the novelty of the sampled graphs by comparing them against the training and validation sets. We use the test set for the calculation of the compression length when training has finished. The best performing models for each dataset are \textbf{bolded}. Baseline results are from the IntelliGraphs paper \citep{thanapalasingam2023intelligraphs}. The full results are available in Tables \ref{graph-sampling-results_appendix} and \ref{compression-results} in the Appendix.
}
\label{tab:graph-sampling-results}
\begin{center}
\small
\begin{tabular}{llcccc}
\thickhline
\multirow{2}{*}{\textbf{Datasets}} & \multirow{2}{*}{\textbf{Model}} & \textbf{\% Valid} & \textbf{\% Novel} & \textbf{\% Empty} & \textbf{Compression} \\
& & \textbf{Graphs} $\uparrow$ & \textbf{\& Valid} $\uparrow$ & \textbf{Graphs} $\downarrow$ & \textbf{Length (bits/graph)} $\downarrow$ \\
\thickhline
\multirow{7}{*}{\textbf{syn-paths}}
& uniform & 0 & 0 & 0 & 30.49 \\
& TransE & 0.25 & 0.25 & 76.55 & 49.89 \\
& DistMult & 0.69 & 0.69 & 85.41 & 54.39 \\
& ComplEx & 0.71 & 0.71 & 85.73 & 48.58\\
& \transformerbaselineabbr & 99.60 & 99.60 & 0 & 27.77 \\
& \modelvae & 92.50 & 92.50 & 0 & 28.74 \\
& \transformerark & 97.39 & 97.39 & 0 & \textbf{27.57} \\
& \modelname  &  \textbf{99.95}& \textbf{99.95}& 0 & 27.65 \\
\hline
\multirow{7}{*}{\textbf{syn-tipr}}
& uniform & 0 & 0 & 0 & 61.61 \\
& TransE & 0 & 0 & 94.42 & 69.51 \\
& DistMult & 0 & 0 & 86.66 & 63.96 \\
& ComplEx & 0 & 0 & 96.05 & 67.51 \\
& \transformerbaselineabbr & 100.00 & 100.00 & 0 & 26.30 \\
& \modelvae & 98.45 & 98.45& 0 & 27.14 \\
& \transformerark & 100.00 & 100.00 & 0 & \textbf{23.34} \\
& \modelname  & \textbf{100.00}  &\textbf{100.00} & 0 & 23.48 \\
\hline
\multirow{7}{*}{\textbf{syn-types}}
& uniform & 0 & 0 & 0 & \textbf{36.02} \\
& TransE & 0.21 & 0.21 & 84.56 & 48.26 \\
& DistMult & 0.13 & 0.13 & 87.53 & 47.69 \\
& ComplEx & 0.07 & 0.07 & 89.75  & 47.69\\
& \transformerbaselineabbr & \textbf{100.00} & \textbf{100.00} & 0 & 59.61 \\
& \modelvae & 100.00 & 100.00 & 0 & 60.58\\
& \transformerark & 87.07 & 87.07 & 0 & 59.79 \\
& \modelname  & 89.22 & 89.22& 0 & 59.63 \\ 
\hline
\multirow{7}{*}{\textbf{wd-movies}}
& uniform & 0 & 0 & 0 & 171.60 \\
& TransE & 0 & 0 & 85.39 & 208.60 \\
& DistMult & 0 & 0 & 87.07& 202.68 \\
& ComplEx & 0 & 0 & 98.13 & 208.50 \\
& \transformerbaselineabbr & \textbf{99.83} & \textbf{99.83} & 0 & 124.50 \\
& \modelvae & 99.47 &99.47 & 0 & 116.84 \\
& \transformerark & 98.33 & 98.33 & 0 & 114.49 \\
& \modelname  & 99.24 & 99.24 & 0 & \textbf{98.19} \\
\hline
\multirow{7}{*}{\textbf{wd-articles}}
& uniform & 0 & 0 & 0 &  693.80 \\
& TransE & 0 & 0 & 95.42 &  910.65 \\
& DistMult & 0 & 0 & 100.00  & 887.30 \\
& ComplEx & 0 & 0 & 97.54 &  901.91 \\
& \transformerbaselineabbr & 98.00 & 98.00  & 0 & 235.24 \\
& \modelvae & \textbf{99.13} & \textbf{99.13 }& 0 & \textbf{199.55} \\
& \transformerark & 95.37 & 95.37 & 0 & 224.25 \\
& \modelname  &97.24  &97.24 & 0 & 205.24 \\
\thickhline
\end{tabular}
\end{center}
\end{table}

We assess the generative capabilities of \modelvae through two complementary approaches: unconditional generation by sampling from the prior distribution $p_\theta(z)$, and conditional generation by providing partial graph sequences. These experiments test whether the learned latent space is well-structured and whether the model can generate semantically valid, novel graphs, demonstrating true generative modeling rather than mere memorization. For more details regarding the method and qualitative analysis, we refer the reader to Appendices \ref{subsec:Sampling_from_latent_variable_z} and \ref{sec:qualitiative-sampling-samplaing}, respectively.

\noindent \textbf{Quantitative Results} \quad Table \ref{tab:graph-sampling-results} shows unconditional graph generation results. \modelname achieves high semantic validity across synthetic datasets: 99.95\% on syn-paths, 100.00\% on syn-tipr, and 89.22\% on syn-types. \modelvae demonstrates similarly strong performance with 92.50\%, 98.45\%, and 100.00\% validity, respectively. Both models dramatically outperform KGE baselines (TransE, DistMult, ComplEx), which achieve less than 1\% validity and produce 76--100\% empty graphs, confirming that independent triple scoring cannot capture the joint structure required for KG generation. All generated graphs from our models are novel rather than memorizing training examples. For real-world datasets, ARK maintains 99.24\% validity on wd-movies and 97.24\% on wd-articles, while \modelvae achieves 99.47\% and 99.13\% respectively, demonstrating robust performance despite increased complexity.

\subsection{Interpolation in Latent Space}

\begin{figure}[!t]
    \centering
    \begin{subfigure}[b]{0.48\textwidth}
        \centering
        \includegraphics[width=\textwidth]{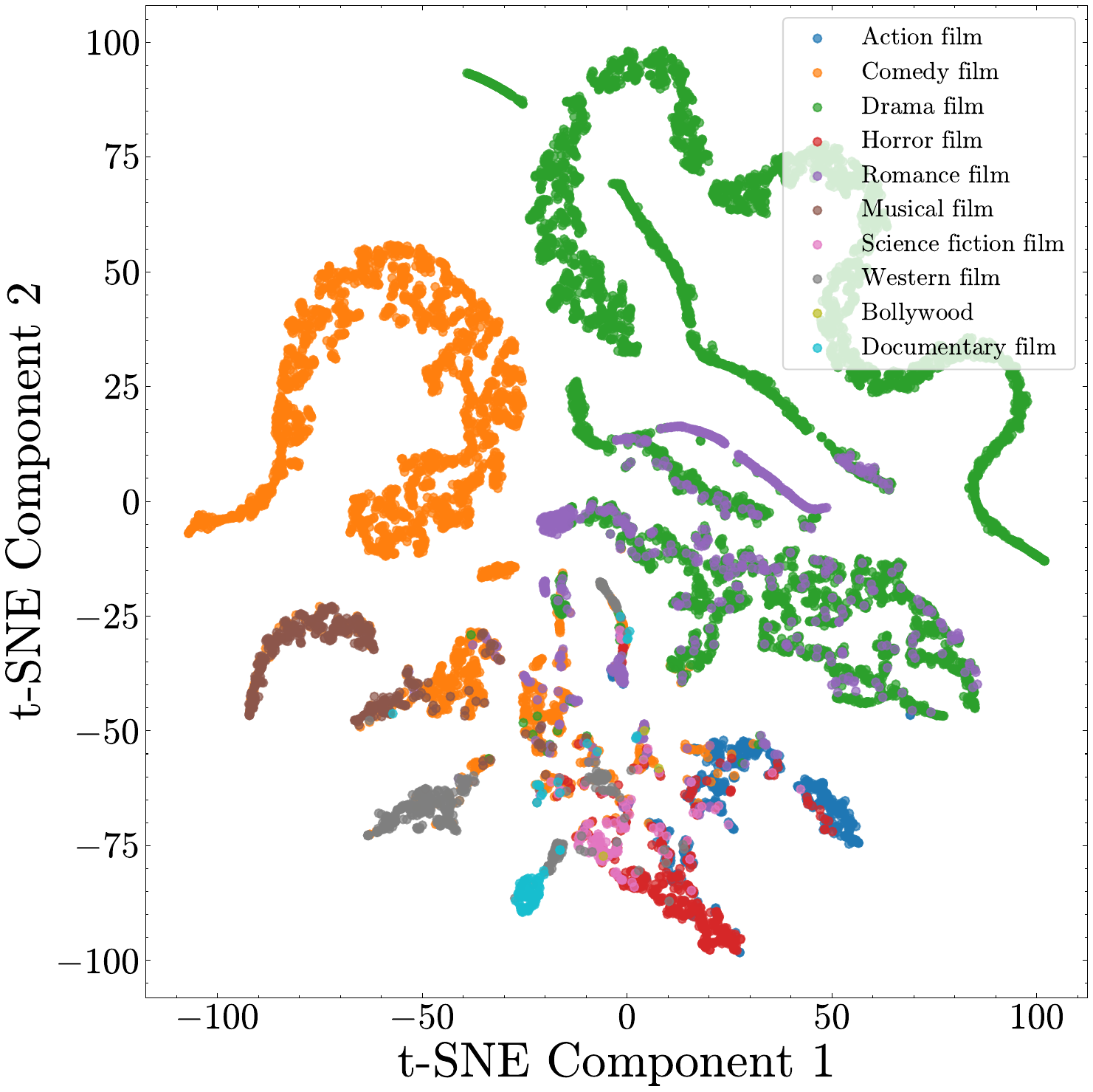}
        \caption{t-SNE projection colored by genre}
        \label{fig:tsne_movies}
    \end{subfigure}
    \hfill
    \begin{subfigure}[b]{0.48\textwidth}
        \centering
        \includegraphics[width=\textwidth]{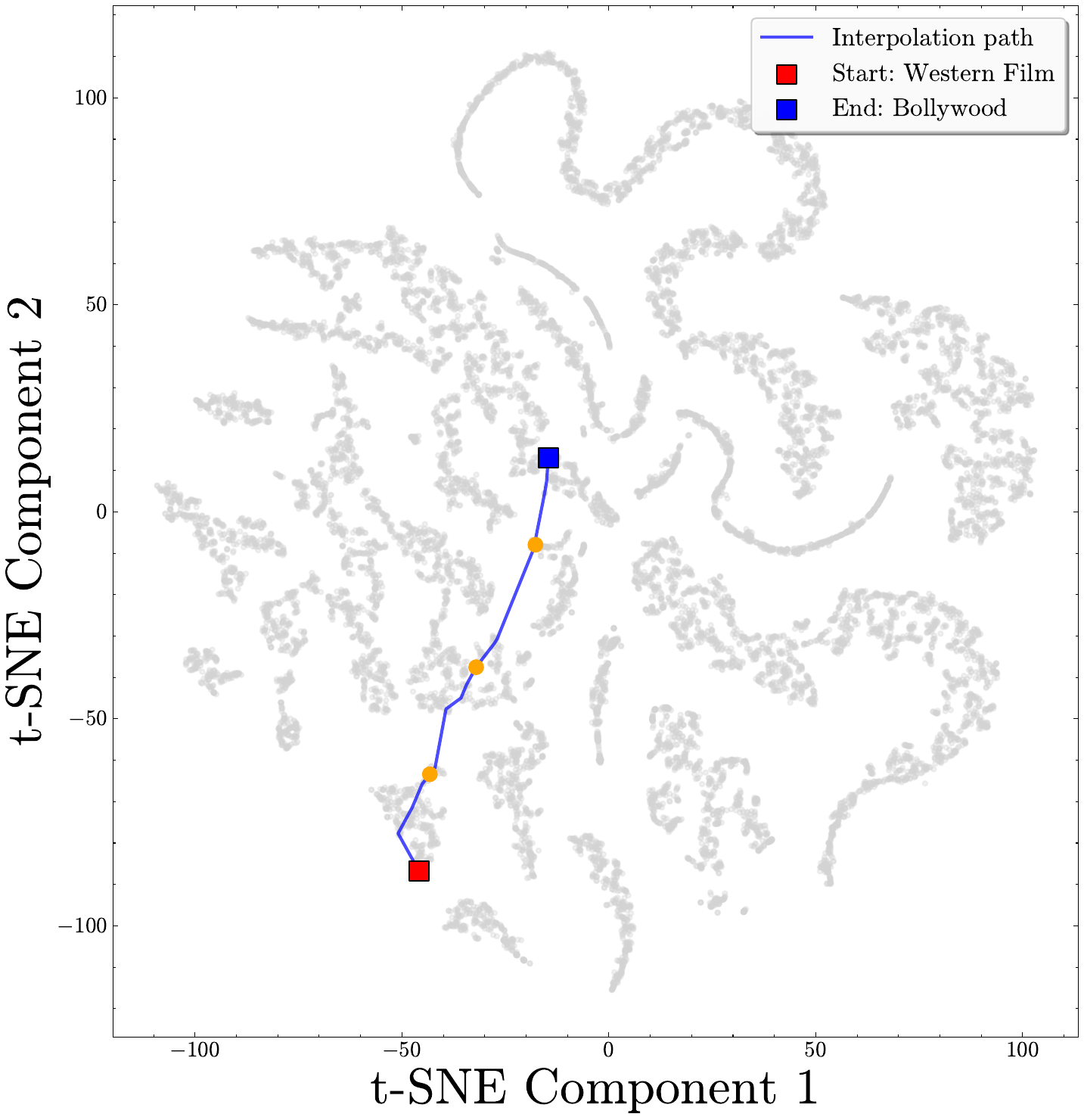}
        \caption{Linear interpolation between two movies}
        \label{fig:interpolation}
    \end{subfigure}
    
    \vspace{0.3cm}
    
    \begin{subfigure}[b]{0.90\textwidth}
        \centering
        \includegraphics[width=\textwidth]{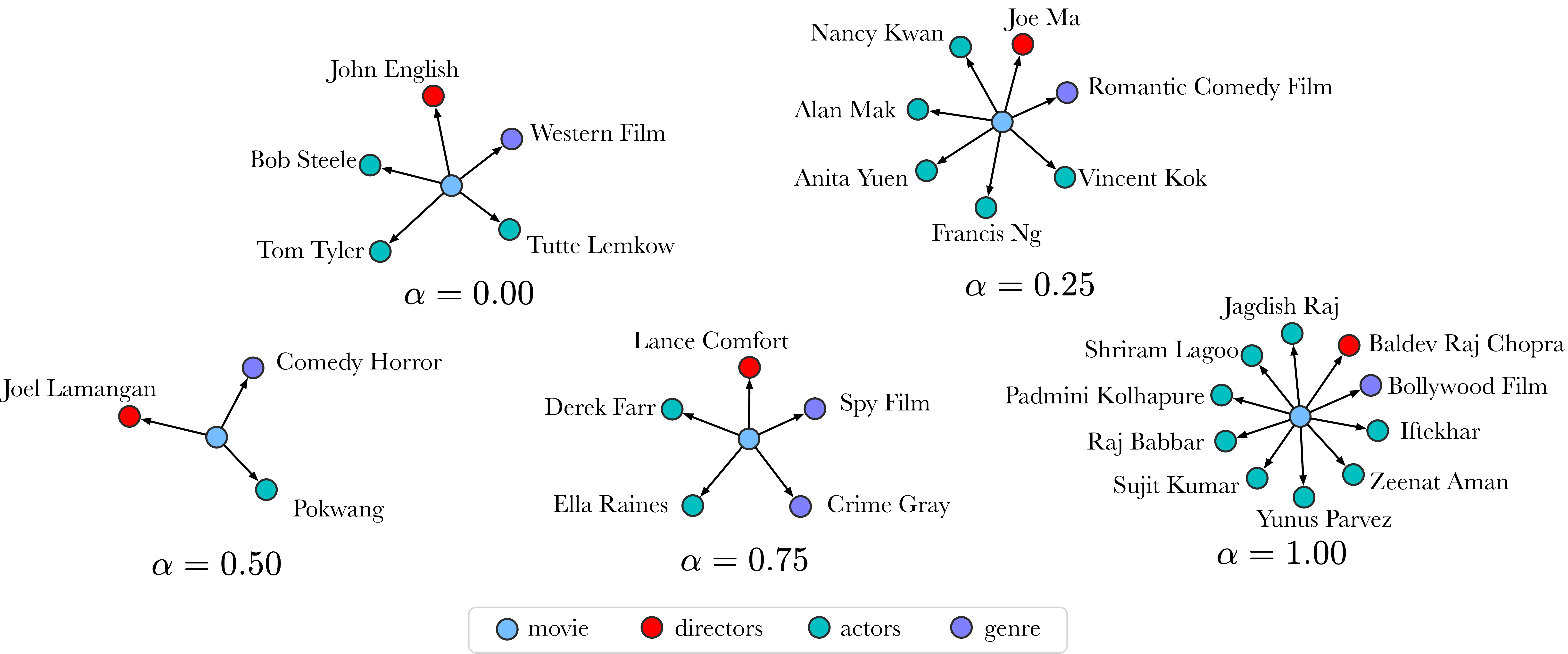}
        \caption{Decoded graphs at interpolation points $\alpha \in \{0.00, 0.25, 0.50, 0.75, 1.00\}$}
        \label{fig:interp_sequence}
    \end{subfigure}
    
    \caption{Latent space visualization for the wd-movies dataset. (a) t-SNE projection shows clear clustering by genre. (b) Smooth interpolation paths connect different movie types. (c) Decoded graphs along the interpolation path show gradual transitions in cast and genre attributes, maintaining semantic validity throughout.}
    \label{fig:latent-visualization}
\end{figure}

For \modelvae and \transformerbaselineabbr, we explore the structure of the learned latent space by interpolating between encoded representations of different graphs. This analysis reveals whether the model learns smooth, semantically meaningful transitions between graph structures, indicating a well-organized latent space where similar graphs cluster together and intermediate points correspond to valid hybrid structures. For more details regarding the method, we refer the reader to Appendix \ref{subsec:interpolation_in_latent_space}.

\noindent \textbf{Quantitative Results} \quad The smoothness metrics reveal distinct patterns across dataset complexity. Comparing \modelvae and \transformerbaselineabbr (Figure \ref{tab:latent_smoothness_results_models} in Appendix), we observe that \transformerbaselineabbr generally achieves better latent space organization. For syn-tipr, \transformerbaselineabbr shows exceptional quality with near-perfect local smoothness (0.99) and global consistency (0.98), while \modelvae achieves lower but still strong metrics (0.93 and 0.69, respectively). The architectural difference is most pronounced on syn-paths, where \transformerbaselineabbr maintains moderate global consistency (0.36) compared to \modelvae's much weaker performance (0.14), suggesting that transformer-based encoders better capture graph structure. Surprisingly, \modelvae demonstrates superior performance on syn-types with high local smoothness (0.92) and global consistency (0.73), exceeding \transformerbaselineabbr's metrics (0.82 and 0.60). The flip rates reveal interesting trade-offs: \modelvae shows higher instability on most datasets (0.33 for syn-paths, 0.40 for wd-movies) compared to \transformerbaselineabbr (0.20 and 0.15 respectively), though they achieve similar rates on syn-tipr (0.10 vs. 0.09). Real-world datasets (wd-movies) show both models achieving strong local smoothness (0.84 and 0.87) but with \transformerbaselineabbr maintaining better global consistency (0.58 vs 0.49). These results suggest that while transformer encoders generally provide better latent space organization, the simpler GRU-based \modelvae can match or exceed transformer performance on certain structured datasets, particularly syn-types.

\noindent \textbf{Qualitative Results} \quad Figure \ref{fig:latent-visualization} demonstrates the learned latent space structure for the wd-movies dataset. The t-SNE projection (Figure \ref{fig:latent-visualization} a) reveals distinct clustering by genre, indicating that \modelvae learns to organize its latent space according to semantic film categories without explicit supervision. The linear interpolation experiment (Figure \ref{fig:latent-visualization} b,c) traces a path between a Western film and a Romantic Comedy, with decoded graphs at intermediate points ($\alpha \in \{0, 0.25, 0.5, 0.75, 1\}$) demonstrating smooth transitions: starting from a Western with actors Bob Steele and Tom Tyler, progressing through hybrid representations with mixed genre elements (Comedy Horror at $\alpha = 0.50$), and reaching a Thriller film with different cast members. While all intermediate graphs maintain a valid KG structure, the semantic coherence varies; intermediate points produce valid but potentially less realistic combinations of actors and genres, suggesting that semantic validity is preserved throughout interpolation, but semantic plausibility is highest near the training data manifold, consistent with typical VAE behavior on structured data.

\subsection{Ablation Study}
The ablation study in Appendix \ref{sec:ablation-study} examines the relationship between model capacity and generation quality. We find that hidden dimensionality is more critical than network depth, with a clear performance threshold at $d_{\text{model}} = 64$. Single-layer GRU models with sufficient capacity ($\geq$ 64 hidden units) match the validity of deeper architectures, suggesting that KG generation does not require complex multi-layer models. Compression efficiency improves with model complexity, yet lightweight architectures match transformer-based models in generation quality.

\section{Related Work}
\label{sec:related-work}
While substantial progress has been made in related areas such as molecular graph generation and graph neural networks, the unique challenges of KG generation, including semantic consistency, relational diversity, and logical constraints, have only begun to be addressed. Recent advances in graph representation learning have paved the way for developing generative modeling of KGs. 

\noindent \textbf{Generative Modeling of KGs} \quad
\citet{cowen2019neural} learn joint probability distributions over facts stored in Knowledge Graphs to estimate the predictive uncertainty of KGE models and evaluate their generative model using link prediction. TransG is a probabilistic model that learns the semantics of N-\emph{ary} relations \citep{xiao-etal-2016-transg}. In contrast to this work, we focus on generating a collection of triples. Loconte \emph{et al.} reinterprets the score functions of traditional KGEs as circuits, enabling efficient marginalization and sampling, thereby facilitating the generation of new triples consistent with existing KGs \citep{loconte2024turn}. Notably, \citet{galkin2024towards} proposes ULTRA, a foundation model for KG reasoning that achieves strong generalization across diverse KGs through a unified pre-training approach on multiple graphs, demonstrating that a single model can transfer reasoning capabilities across different knowledge domains without fine-tuning. Work focusing on triple completion from various sources includes: \citet{zhang2024startfromzero} which addresses triple set prediction, and \citet{sun2018bootea} which focuses on entity alignment. \citet{neelakantan-etal-2015-compositional} uses RNNs for scoring paths in existing graphs (link prediction), whereas we generate complete novel graphs from learned distributions—a fundamentally different task.

\textbf{Graph Transformers} \quad
Machine learning on sets requires learning permutation-invariant functions \citep{zaheer2017deep}. Various frameworks have been proposed that use attention mechanisms for graph representation learning \citep{kim2022pure, yun2019graph, zhuo2025dualformer, zhaographgpt}. Due to the fully-attentional nature of Transformers, they can be seen as a generalisation of Graph Neural Networks \citep{zaheer2017deep}. In our work, we deal with a directed graph with labeled edges. Recent advances include GraphGPS \citep{rampavsek2022recipe}, which combines message passing with global attention mechanisms, and NodeFormer \citep{wu2022nodeformer}, which efficiently computes all-pair interactions through kernelized softmax. \citet{shirzad2023exphormer} introduce Exphormer, achieving linear complexity in graph transformers through virtual global nodes and expander graphs. Unlike these architectures that focus on encoding existing graphs, our work addresses the complementary problem of generating new KGs through autoregressive sequence modeling.

\textbf{Graph Generative Models} \quad
Deep graph generative models have predominantly focused on generating novel molecular structures, emphasizing chemical validity and stability \citep{li2018learning}. Beyond molecular applications, models like GraphVAE and GraphRNN have been developed to capture complex graph structures through latent variable and autoregressive approaches. \cite{kipf2019contrastive} introduced methods to infer symbolic abstractions from visual data and relational structures from observations. Recent developments include DiGress \citep{vignac2023digress}, which applies discrete denoising diffusion to graph generation, and GraphARM \citep{kong2023autoregressive}, which combines autoregressive models with graph neural networks for scalable generation. \citet{liu2024graphmaker} propose GraphMaker, a diffusion-based approach that generates graphs by iteratively refining node features and edge structures. Our work extends these concepts by focusing on the semantic generation of KGs, learning implicit semantic constraints from background information without predefined rules.

\textbf{Neuro-Symbolic Generative Models for KGs} \quad
Combining distributed and symbolic representations, neuro-symbolic systems aim to combine the strengths of both paradigms \citep{van2021modular}. Generative neuro-symbolic machine combines distributed and symbolic entity-based representations in a generative latent variable model to infer object-centric symbolic representations from images \citet{jiang2020generative}. \cite{balloccu2024kgglm} introduces KGGLM, a generative language model designed for generalizable KG representation learning in recommender systems, which exemplifies the integration of neural and symbolic approaches. A recent breakthrough comes from \citet{vankrieken2025neurosymbolic}, who introduce a neurosymbolic diffusion model that integrates logical constraints directly into the diffusion process. While these approaches explicitly incorporate symbolic reasoning into neural networks, our work demonstrates that autoregressive models can implicitly learn semantic constraints from data. 

\section{Conclusion}
\label{sec:conclusion}

We have demonstrated that autoregressive sequence models provide an effective framework for Knowledge Graph generation. By representing KGs as sequences of (head, relation, tail) triples, our models learn to generate semantically valid graphs that satisfy domain constraints without explicit rule supervision. \modelname and \modelvae achieve 89.2\% to 100.0\% validity across diverse datasets, substantially outperforming KGE baselines that treat triples independently. The variational extension \modelvae further enables controlled generation through learned latent representations, supporting both unconditional sampling and conditional completion from partial graphs. Our analysis reveals that hidden dimensionality matters more than architectural depth for this task, with single-layer GRU models matching the validity of deeper transformer architectures while offering computational efficiency. Our findings establish that autoregressive models can effectively learn the joint distribution over KG structures, implicitly capturing semantic rules. This opens avenues for KG completion, data augmentation, and query answering applications where generating valid local structures is the primary requirement.

\noindent \textbf{Limitations} \quad Our work assumes a fixed vocabulary of entities and relations known at training time, limiting applicability to open-world scenarios where new entities emerge dynamically. While \modelname and \modelvae generate semantically valid graphs, in our experiments, we only test on relatively small Knowledge Graphs (3-212 triples). We focus on learning valid compositional patterns of known elements. While this limits applications requiring truly open-world generation, many practical use cases (KG completion, query answering over existing KGs) naturally operate under these assumptions. Moreover, our learned compositional patterns could potentially transfer to new entities through inductive embeddings in future work. IntelliGraphs focuses on subgraph generation rather than complete KG synthesis, which aligns with real-world applications that typically require coherent subgraphs. Additionally, the autoregressive formulation imposes a linear ordering on inherently unordered graph structures, though our experiments show that this does not significantly impact generation quality.

\noindent \textbf{Future Work} \quad Several directions merit exploration: extending the \modelname framework to handle \emph{out-of-vocabulary} entities and relations through compositional embeddings or meta-learning approaches, investigating hierarchical generation strategies for larger graphs where local subgraphs are generated independently then composed, and making the learned semantic rules explicit rather than leaving them implicit in the model parameter would help to identify and mitigate learning undesired constraints that may stem from biases in the data. Since this work solves most challenges in the Intelligraphs benchmark, larger and more complex KG generation benchmarks are called for. Future directions include: (1) extending to open-world generation with compositional embeddings or meta-learning for handling new entities/relations dynamically, (2) scaling beyond 500 triples through hierarchical generation strategies (generating subgraphs then composing them), (3) integration with LLMs to combine our structured generation with natural language understanding, (4) making learned semantic rules explicit for better interpretability and bias detection, and (5) developing larger, more complex benchmarks since our models solve most IntelliGraphs challenges.

\newpage

\noindent \textbf{Ethics Statement} \quad
Datasets on which our models are trained may contain societal biases and factual errors, which could propagate through the learning process and manifest in generated knowledge graphs. While our models achieve high semantic validity scores, they may still reproduce or amplify biases present in the training data, potentially generating graphs that reflect historical inequities or stereotypes. Additionally, the autoregressive generation process could produce factually incorrect but semantically valid triples, as the model learns logical rules rather than verifying the truth. We intend for \modelname and \modelvae to be treated as research prototypes to advance the field of KG generation, and should not be deployed in critical applications without thorough testing and safeguards. See \citet{thanapalasingam2023intelligraphs} for a detailed analysis of the limitations of the datasets.

\noindent \textbf{Reproducibility Statement} \quad
We provide complete code and detailed configurations to ensure complete reproducibility of all experiments. Our implementation, including model architectures, training scripts, data preprocessing pipelines, and evaluation metrics, is available at \repolink. Experimental details, including hyperparameters, hardware specifications, and training procedures, are provided in Appendix \ref{sec:experimental-details}. We also release pre-trained model checkpoints for both \modelname and \modelvae to facilitate reproduction of our results and enable further research building upon our work. Detailed instructions for replicating each experiment, including expected runtimes and resource requirements, are provided in the repository.

\bibliography{main}
\bibliographystyle{tmlr}

\newpage

\appendix
\section{Appendix} 

\subsection{Experimental Details}
\label{sec:experimental-details}

We used the PyTorch library \footnote{\url{https://pytorch.org/}} to develop and test the models. All experiments were performed on a single-node machine with an Intel(R) Xeon(R) Gold 5118 (2.30GHz, 12 cores) CPU and 64GB of RAM, with a single NVIDIA A100 GPU (80GB of VRAM) or a single NVIDIA H100 GPU (80GB of VRAM). We used PyTorch's CUDA acceleration for model training and inference. We used the Adam optimizer with variable learning rates \citep{kingma2014adam}. We monitored the training of the models using the Weights \& Biases package \footnote{\url{https://wandb.ai}}. All experiments use the same train/validation/test splits as the original IntelliGraphs benchmark \citep{thanapalasingam2023intelligraphs} to ensure fair comparison.

\textbf{Hyperparameter Optimization} \quad
For \modelname and \modelvae, the hyperparameters were automatically tuned using grid search \{learning rate, batch size, number of epochs, latent dimension size \footnote{for \modelvae only}, number of neurons and number of layers\footnote{Both encoder's and decoder's neurons and number of layers. For models without encoder the tuning for the number of layers and neurons was done the decoder part} \} to get the best performance for the validation split. For reproducibility, we provide an extension description of the hyperparameters as YAML files under the \emph{configs} directory on \repolink. 

\subsection{Dataset Details}
\label{app:dataset-details}

The IntelliGraphs benchmark datasets test different aspects of semantic validity and structural complexity:

\begin{enumerate}
    \item \textbf{syn-paths:} A synthetic dataset containing path graphs with simple semantics that can be algorithmically verified in linear time. These are acyclic graphs where edge directions follow the path structure.
    
    \item \textbf{syn-types:} A synthetic dataset featuring typed entities and relations where type constraints on entities depend on the relation type, enforcing semantic consistency through type checking.
    
    \item \textbf{syn-tipr:} A synthetic dataset containing subgraphs based on the \emph{Time-indexed Person Role} (tipr) ontology pattern.\footnote{http://ontologydesignpatterns.org/wiki/Submissions:Time\_indexed\_person\_role} The semantics are defined by the tipr graph pattern, requiring temporal reasoning to generate valid time intervals.
    
    \item \textbf{wd-movies:} Small knowledge graphs describing movies, extracted from Wikidata.\footnote{https://www.wikidata.org} Each graph contains one existential node representing the movie, with entity nodes for director(s) connected via \texttt{has\_director}, cast members connected via \texttt{has\_actor}, and genres connected via \texttt{has\_genre} relations.
    
    \item \textbf{wd-articles:} Small knowledge graphs that describe research articles, extracted from Wikidata. Each graph contains one existential node representing the article, with entity nodes for author(s) connected via \texttt{has\_author}, publication venues connected via \texttt{published\_in}, and topics connected via \texttt{has\_topic} relations.
\end{enumerate}

\begin{table}[H]
\caption{Dataset characteristics for the IntelliGraphs benchmark. Synthetic datasets (syn-*) have fixed graph structures while Wikidata-derived datasets (wd-*) exhibit variable sizes. Entity counts represent unique entities across all graphs; edge counts indicate the number of triples per individual graph.}
\label{tab:dataset-stats-appendix}
\begin{center}
\begin{tabular*}{\textwidth}{@{\extracolsep{\fill}}lcccc@{}}
\thickhline
\multirow{2}{*}{\textbf{Datasets}} & \textbf{Dataset Size} & \textbf{Unique} & \textbf{Relation} & \textbf{Triples per}\\
 & \textbf{(Train/Val/Test)} & \textbf{Entities} & \textbf{Types} & \textbf{Graph}\\
\thickhline
syn-paths & 60,000/20,000/20,000 & 49 & 3 & 3\\
syn-types & 60,000/20,000/20,000 & 30 & 3 & 3\\
syn-tipr & 50,000/10,000/10,000 & 130 & 5 & 5\\
wd-movies & 38,267/15,698/15,796 & 24,093 & 3 & 2-23\\
wd-articles & 54,163/22,922/22,915 & 60,932 & 6 & 4-212\\
\thickhline
\end{tabular*}
\end{center}
\end{table}

\subsection{Methods}
\label{sec:methods}

Here, we provide more details about the methods we used for the empirical analyses of \modelname and \modelvae.

\subsubsection{Gated Recurrent Units (GRUs)}
\label{subsec:gru-details}

The \modelname model employs a standard GRU decoder with hidden state $\mathbf{h}_t \in \mathbb{R}^d$ that evolves as:
\begin{align}
\mathbf{r}_t &= \sigma(\mathbf{W}_r \mathbf{x}_t + \mathbf{U}_r \mathbf{h}_{t-1} + \mathbf{b}_r) \\
\mathbf{z}_t &= \sigma(\mathbf{W}_z \mathbf{x}_t + \mathbf{U}_z \mathbf{h}_{t-1} + \mathbf{b}_z) \\
\tilde{\mathbf{h}}_t &= \tanh(\mathbf{W}_h \mathbf{x}_t + \mathbf{U}_h (\mathbf{r}_t \odot \mathbf{h}_{t-1}) + \mathbf{b}_h) \\
\mathbf{h}_t &= (1 - \mathbf{z}_t) \odot \mathbf{h}_{t-1} + \mathbf{z}_t \odot \tilde{\mathbf{h}}_t
\end{align}

where $\mathbf{r}_t$ and $\mathbf{z}_t$ are reset and update gates respectively, $\mathbf{x}_t$ is the embedding of the current input token, and $\odot$ denotes element-wise multiplication. At each timestep, the hidden state is projected to vocabulary logits:
$p(x_{t+1} | x_{\leq t}) = \text{softmax}(\mathbf{W}_o \mathbf{h}_t + \mathbf{b}_o)$. 

\subsubsection{Compression Length}
\label{subsec:compression_length}

For both \modelname and \modelvae, we compute the compression length to generate graphs as sequences. Since \modelname is a decoder-only autoregressive model, we compute:

\begin{align}
\text{Compression Length of } G = -\log_2(p_\theta(G)) = -\sum_{t=1}^{T} \log_2(p_\theta(x_t | x_{<t}))
\end{align}

where $x_t$ represents the $t$-th token in the linearized graph sequence $[\texttt{BOS}, h_1, r_1, t_1, ..., \texttt{EOS}]$ and $T$ is the sequence length. Each term represents the bits needed to encode the next token given the previous context.

For \modelvae, the variational framework adds a latent variable $z$, resulting in an upper bound on compression length through the ELBO:

\begin{align}
\label{eq:compression-spark}
\text{Compression Length of } G &\leq -\log_2(p(G | z)) + D_{\text{KL}}( q(z\mid G) \parallel p(z) ) \\
&= -\sum_{t=1}^{T} \log_2(p_\theta(x_t | x_{<t}, z)) + D_{\text{KL}}
\end{align}

The KL divergence term is computed as follows:
\begin{equation}
D_{\text{KL}}( q(z\mid G) \parallel p(z) ) = \frac{1}{2} \sum_{i=1}^{d} \left( \mu_i^2 + \sigma_i^2 - 1 - \log(\sigma_i^2) \right) \cdot \log_2(e)
\end{equation}

where $d$ is the latent dimensionality and the factor $\log_2(e)$ converts from nats to bits. The autoregressive formulation naturally handles variable-length graphs through the sequential factorization, eliminating the need for separate structure and entity terms.

This provides an upper bound on the true compression length; the VAE's ELBO is a lower bound on log-likelihood, which, when negated, becomes an upper bound on compression. The bound is particularly relevant as the autoregressive decoder must account for uncertainty in token ordering during generation.

\subsubsection{Sampling from Latent Variable, $z$}
\label{subsec:Sampling_from_latent_variable_z}
We conduct two types of generation experiments:
\begin{enumerate}
    \item \emph{Unconditional Generation:} We sample 10,000 random latent codes from the standard normal prior distribution $p(z) = \mathcal{N}(0, I)$ and decode them into complete graphs using beam search with beam width $k=3$. Each decoded graph is analyzed for: (1) semantic validity according to dataset-specific constraints, (2) novelty by checking against the training and validation sets, and (3) non-emptiness to ensure the model generates meaningful structures rather than null graphs.
    \item \emph{Conditional Generation:} We evaluate the model's ability to complete partial graphs by providing incomplete sequences as prompts. For each test graph, we provide the first $n$ tokens (\textit{e.g.}, $[\texttt{BOS}, h_1, r_1, t_1]$) and generate the remaining sequence autoregressively. We vary the conditioning length and measure: (1) the semantic validity of the completed graph and (2) the diversity of completions when sampling with different random seeds.
\end{enumerate}

\subsubsection{Interpolation in Latent Space}
\label{subsec:interpolation_in_latent_space}

We conduct both quantitative and qualitative analyses of the latent space structure:

\begin{enumerate}
    \item \textit{Quantitative Analysis:} We measure latent space smoothness through four metrics: (1) \textit{Local Smoothness} -- average Jaccard similarity between consecutive decoded graphs along random walks in latent space with step size $\epsilon =0.1 $, measuring whether small movements produce similar graphs; (2) \textit{Global Consistency} -- Jaccard similarity between each step and the anchor point, measuring drift from the starting graph; (3) \textit{Flip Rate} -- fraction of steps that produce different decoded graphs, with lower rates indicating larger basins of attraction in latent space; and (4) \textit{Average Basin Length} -- mean number of consecutive interpolation steps that decode to identical graphs, quantifying the granularity of the learned representation. For each metric, we sample multiple anchor points and random directions, taking 10-30 steps along each trajectory. 
    \item \textit{Qualitative Analysis:} We visualize the latent space structure using two approaches: (1) \textit{2D Projection} -- we encode all test graphs and project their latent representations to 2D using t-SNE, coloring points by semantic attributes (genre for wd-movies) to observe clustering patterns; and (2) \textit{Linear Interpolation} -- we select pairs of semantically distinct graphs, encode them to obtain $z_1$ and $z_2$, then decode intermediate points $z_\alpha = (1-\alpha)z_1 + \alpha z_2$ for $\alpha \in [0, 1]$ at regular intervals to examine the semantic coherence of interpolated graphs.
\end{enumerate}

\subsection{Qualitative Analysis of Conditional Sampling}
\label{sec:qualitiative-sampling-samplaing}

\begin{figure}[!h]
    \centering
    \includegraphics[width=0.9\textwidth]{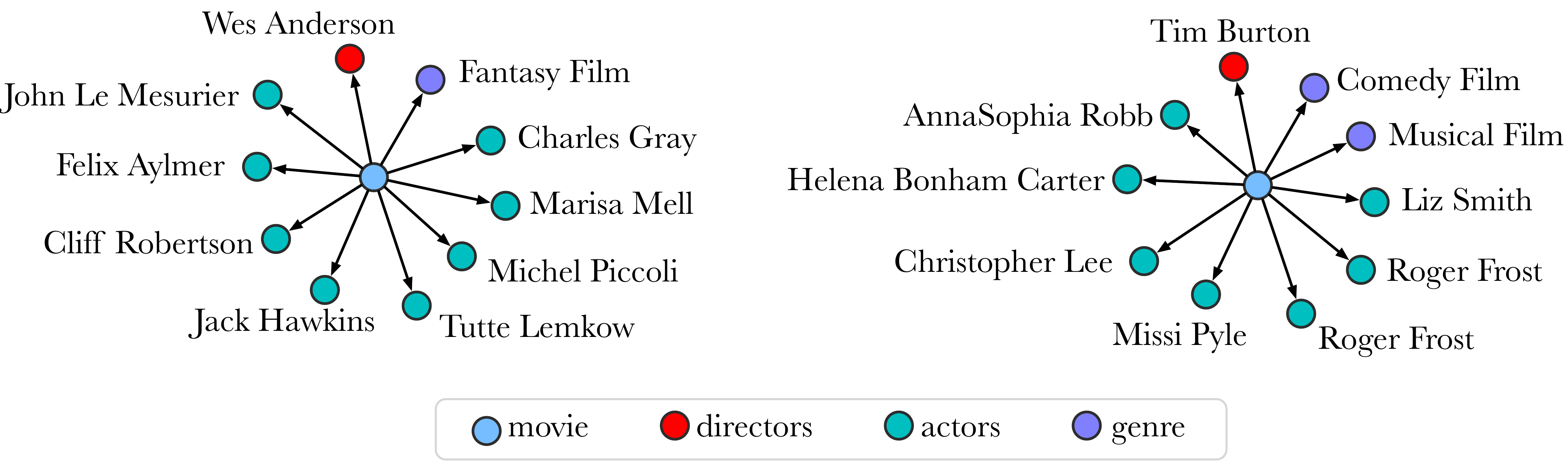}
    \caption{Graphs generated by \modelname conditioned on director entities for Wes Anderson \textbf{(left)} and (b) Tim Burton \textbf{(right)}. Node colors indicate entity types: movie (blue), directors (red), actors (green), and genres (purple).}
    \label{fig:director_generation_comparison}
\end{figure}

\noindent \textbf{Qualitative Results} \quad We test whether \modelvae has learned meaningful latent representations that capture director-specific collaborative patterns and genre preferences, despite never being explicitly trained on individual directorial styles. Figure \ref{fig:director_generation_comparison} shows representative examples of conditional generation for director-specific movie graphs. When conditioned on \quotes{Tim Burton} as the director, the model successfully generates graphs featuring his frequent collaborators (Helena Bonham Carter, Christopher Lee) and characteristic genres (Comedy Film, Musical Film). \modelvae captures Burton's tendency to work repeatedly with the same ensemble cast, demonstrating learned patterns of directorial collaboration. In contrast, the Wes Anderson generation fails to capture his distinctive style. This disparity in generation quality likely reflects differences in dataset representation; Burton's more frequent appearances and consistent casting patterns in the training data enabled better pattern learning, while Anderson's style may have been underrepresented. Despite these variations in director-specific accuracy, both generated graphs maintain semantic validity as movie KGs, indicating that the \modelvae has learned general graph structure.

\subsection{Ablation Study}
\label{sec:ablation-study}

We systematically analyze the contribution of key architectural components through two ablation experiments on the syn-paths dataset, examining both model capacity and architectural choices. 

\textbf{Method} \quad We conduct two complementary ablation studies: 

\begin{enumerate}
    \item \textit{Architectural Hyperparameter Analysis:} We vary the number of GRU layers $n_{\text{layers}} \in \{1, 2, 3, 4, 5\}$ and model dimensions $d_{\text{model}} \in \{2, 4, 8, 16, 32, 64, 128, 256, 512\}$ while keeping other hyperparameters fixed. For each configuration, we train the model until convergence and evaluate generation by measuring the percentage of semantically valid and novel graphs. We also test the relative importance of network depth versus hidden dimensionality on generation quality.
    \item \textit{Architecture Ablation:} We systematically replace transformer components with simpler architectures to assess their contribution: (1) \textit{MLP Encoder} -- replaces the transformer encoder with a multi-layer perceptron while preserving positional encoding; (2) \textit{GRU Decoder} -- replaces the transformer decoder with a GRU-based sequential decoder; and (3) \textit{MLP Encoder \& GRU Decoder} -- combines both modifications, using an MLP encoder and GRU decoder. Each variant maintains comparable parameter counts to the transformer baseline for fair comparison. 
\end{enumerate}

\begin{figure}[h]
    \centering
    \includegraphics[width=\textwidth]{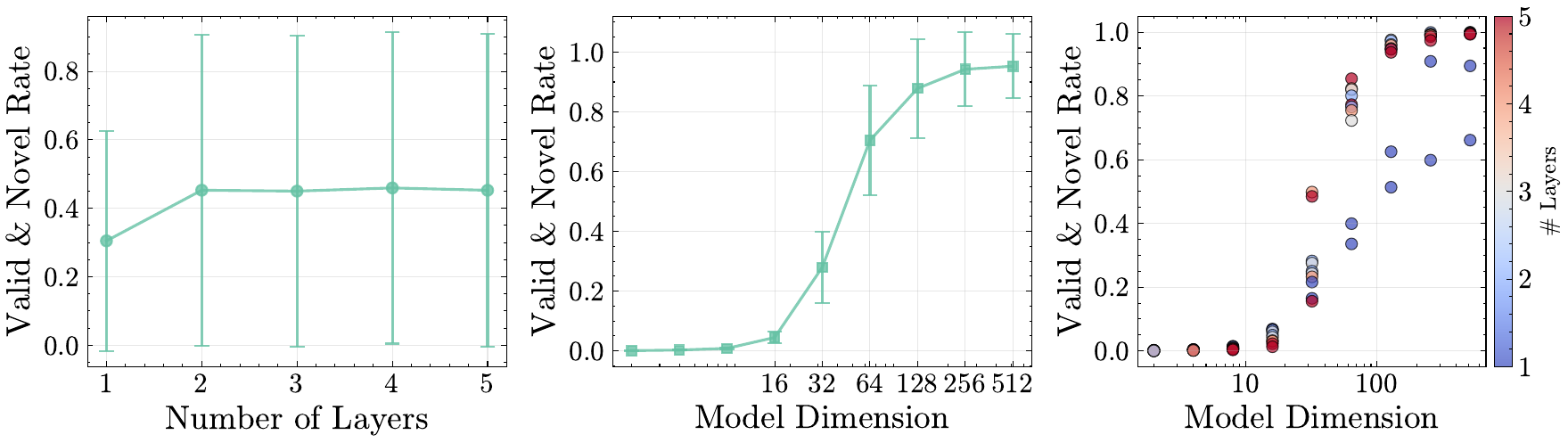}
    \caption{Effect of architectural hyperparameters on the semantic validity and novelty. \textbf{(Left)} Valid \& Novel rate as a function of the number of GRU layers, showing stable performance across depths with high variance. \textbf{(Center)} Performance variation with model dimension (hidden units), demonstrating a sharp improvement threshold around 64 dimensions, followed by consistent high performance. \textbf{(Right)} Scatter plot of individual experimental runs showing the relationship between model dimension and generation quality, with color indicating the number of layers.}
    \label{fig:hyperparameter-analysis}
\end{figure}

\noindent \textbf{Architectural Hyperparameter Analysis Results} \quad In Figure \ref{fig:hyperparameter-analysis}, the model dimension has a substantially stronger impact on generation quality than network depth. Varying the number of layers from 1 to 5 produces relatively stable performance around 45\% valid \& novel rate, though with high variance across configurations. In contrast, the center panel demonstrates a sharp performance threshold: models with fewer than 16 hidden units achieve near-zero validity rates, while those with $d_{\text{model}} \geq 64$ consistently achieve 70-95\% validity. The right panel's scatter plot confirms this pattern across individual runs, showing clear stratification by model dimension rather than layer count (indicated by color). These findings suggest that for KG generation on syn-paths dataset, a single-layer GRU with sufficient hidden units ($\geq$64) can match or exceed the performance of deeper networks, supporting our claim that architectural simplicity does not compromise generation quality when coupled with appropriate capacity.

\label{fig:architectural_hyperparameter}

\noindent \textbf{Architecture Ablation Results} To better understand the contribution of architectural choices, we compare our full transformer-based model \transformerbaselineabbr against simplified variants: \modelvae, which replaces the transformer encoder and decoder with an MLP encoder and a GRU decoder, an MLP encoder (paired with a transformer decoder), \transformerark, a decoder-only transformer model, and \modelname, a GRU decoder-only model. These ablations allow us to isolate the effect of transformer components in both the encoder and the decoder, and to assess whether an encoder is required for KG generation at all. In addition to generation quality and compression efficiency, we also report relative training time, as computational efficiency is often a limiting factor in scaling generative models. \quad Table \ref{tab:architecture-ablation} demonstrates that transformer components, while improving generation quality, are not strictly necessary for effective knowledge graph modeling. Sequential decoders are consistently the most efficient: \modelname trains at \textbf{0.09–0.27},$\times$ the baseline time (i.e., \textbf{3.7–11×} faster) with near baseline validity across datasets, and its sequential inductive bias is competitive for decoding \emph{e.g.}, syn-tipr (23.48 bits, on par with \transformerark’s 23.34) and wd-movies (\textbf{98.19} bits, best overall). Meanwhile, \modelvae yields the best compression on wd-articles ( \textbf{199.55} bits), indicating that modest latent structure plus a GRU decoder can improve efficiency on complex, real-world graphs. Taken together, these results suggest that, for KG generation, a strong sequential decoder often dominates architectural choice, and the extra cost of full transformers, especially in the decoder, may be hard to justify when compute is constrained.

\begin{table}[!h]
\caption{Architectural ablation study comparing \modelname against simplified architectures with MLP encoders and GRU decoders. We evaluate model variants across five datasets using generation quality metrics (percentage of valid and novel graphs), compression efficiency (bits required for latent representation), and computational efficiency (training time relative to \transformerbaselineabbr baseline).}
\label{tab:architecture-ablation}
\begin{center}
\begin{tabular}{llcccc}
\thickhline
\multirow{2}{*}{\textbf{Datasets}} & \multirow{2}{*}{\textbf{Model}} & \textbf{\% Valid} & \textbf{\% Novel} & \textbf{Compression} & \textbf{Training} \\
& & \textbf{Generation} $\uparrow$ & \textbf{Graphs} $\uparrow$ & \textbf{(bits)} $\downarrow$ & \textbf{Time} $\downarrow$ \\
\thickhline 
\multirow{4}{*}{\textbf{syn-paths}}
& \transformerbaselineabbr & 99.60 & 100.00 & 27.77 & 1.00 \\
& \modelvae & 92.50 & 100.00 & 28.74 & 0.21 \\
& MLP Encoder & 99.80 & 100.00 & 27.35 & 0.55 \\
& \transformerark & 97.39 & 100.00 & 27.57 & 0.12  \\
& \modelname & 99.95 & 100.00 & 27.65 & 0.09 \\
\hline
\multirow{4}{*}{\textbf{syn-tipr}}
& \transformerbaselineabbr & 100.00 & 100.00 & 26.30 & 1.00 \\
& \modelvae & 98.45 & 100.00 & 27.14 & 0.17 \\
& MLP Encoder &99.48 &100.00 &26.30 & 0.20\\
& \transformerark & 100.00  & 100.00 & 23.34 & 0.17\\
& \modelname &100 & 100.00 & 23.48 & 0.09\\
\hline
\multirow{4}{*}{\textbf{syn-types}}
& \transformerbaselineabbr & 100.00 & 100.00 & 59.61 & 1.00 \\
& \modelvae & 100.00 & 100.00 & 60.58 & 0.39 \\
& MLP Encoder &93.27 &100.00 &59.33 & 0.41\\
& \transformerark & 87.07  & 100.00 & 59.79 & 0.18 \\
& \modelname & 89.22 & 100.00 & 59.63& 0.09\\
\hline
\multirow{4}{*}{\textbf{wd-movies}}
& \transformerbaselineabbr & 99.83 & 100.00 & 124.50 & 1.00 \\
& \modelvae & 99.47 & 100.00 & 116.84 & 0.24 \\
& MLP Encoder & 99.44 & 100.00 & 118.64 & 0.36 \\
& \transformerark & 98.33 & 100.00 & 114.49 & 0.23 \\
& \modelname & 99.24 & 100.00 &98.19 & 0.21\\
\hline
\multirow{4}{*}{\textbf{wd-articles}}
& \transformerbaselineabbr & 98.00 & 96.00 & 235.24 & 1.00 \\
& \modelvae & 99.13 & 100.00 & 199.55 & 0.42\\
& MLP Encoder & 97.7& 100.00&206.23&0.48\\
& \transformerark & 95.37 & 100.00 & 224.25 & 0.33 \\
& \modelname &97.24 &100.00 & 205.24& 0.27\\
\thickhline
\end{tabular}
\end{center}
\end{table}

\subsection{Conditioned Generation}
\label{subsec:conditioned_gen}

\begin{figure}[htpb]
    \centering
    \begin{subfigure}[b]{0.48\textwidth}
        \centering
        \includegraphics[width=\textwidth]{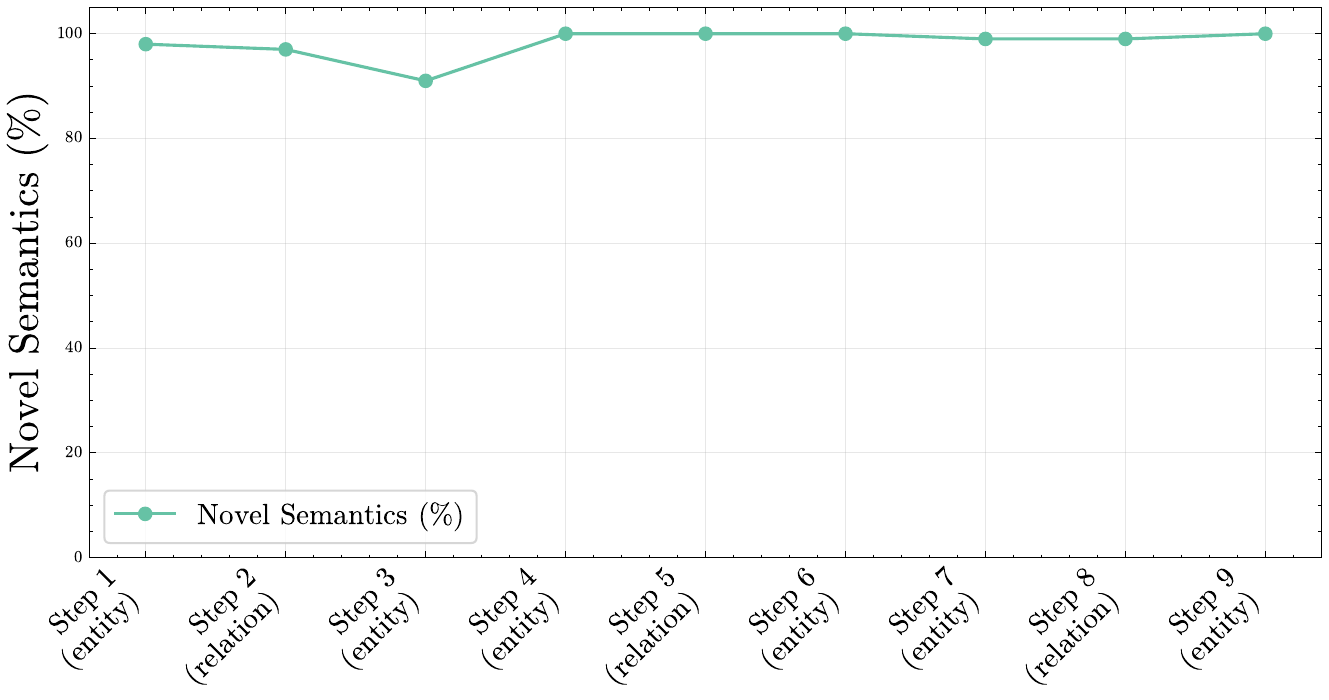}
        \caption{Stepwise conditioning for novelty and validity.}
        \label{fig:novel_constrain}
    \end{subfigure} 
    \begin{subfigure}[b]{0.48\textwidth}
        \centering
        \includegraphics[width=\textwidth]{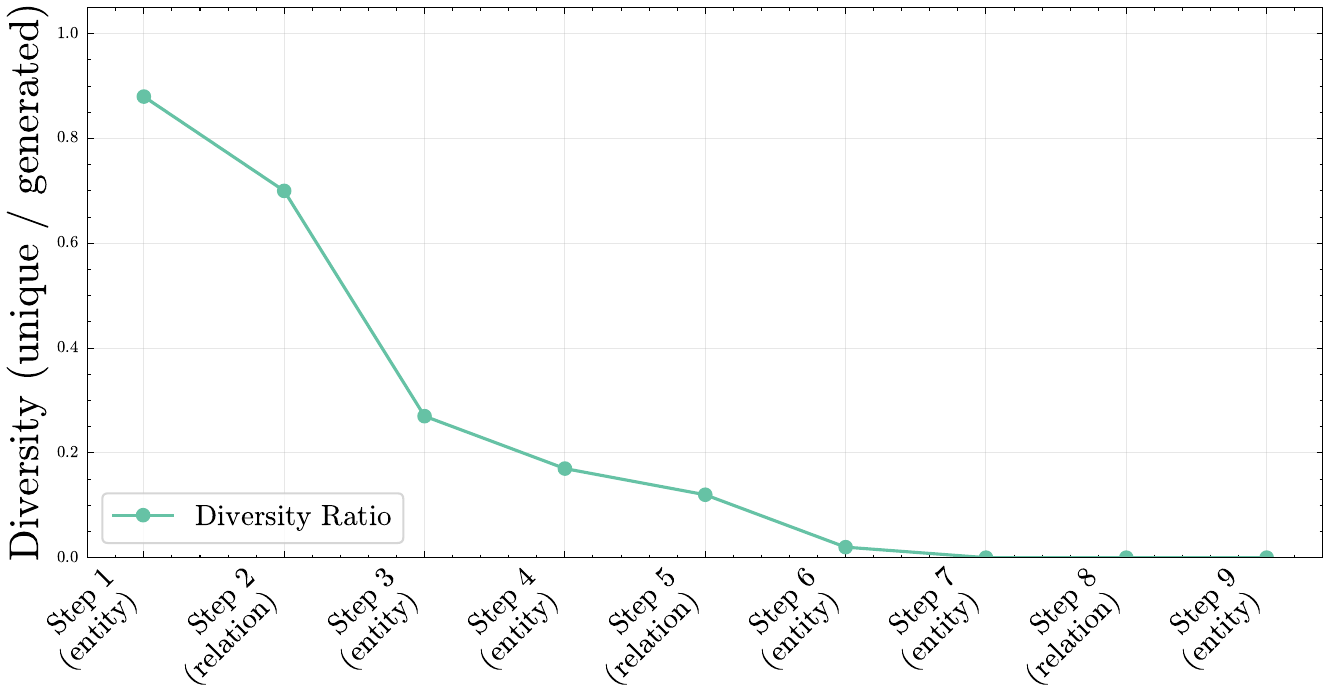}
        \caption{Effect of conditioning on diversity.}
        \label{fig:diversity_constrain}
    \end{subfigure} 
    \begin{subfigure}[b]{\textwidth}
        \centering
        \includegraphics[width=\textwidth]{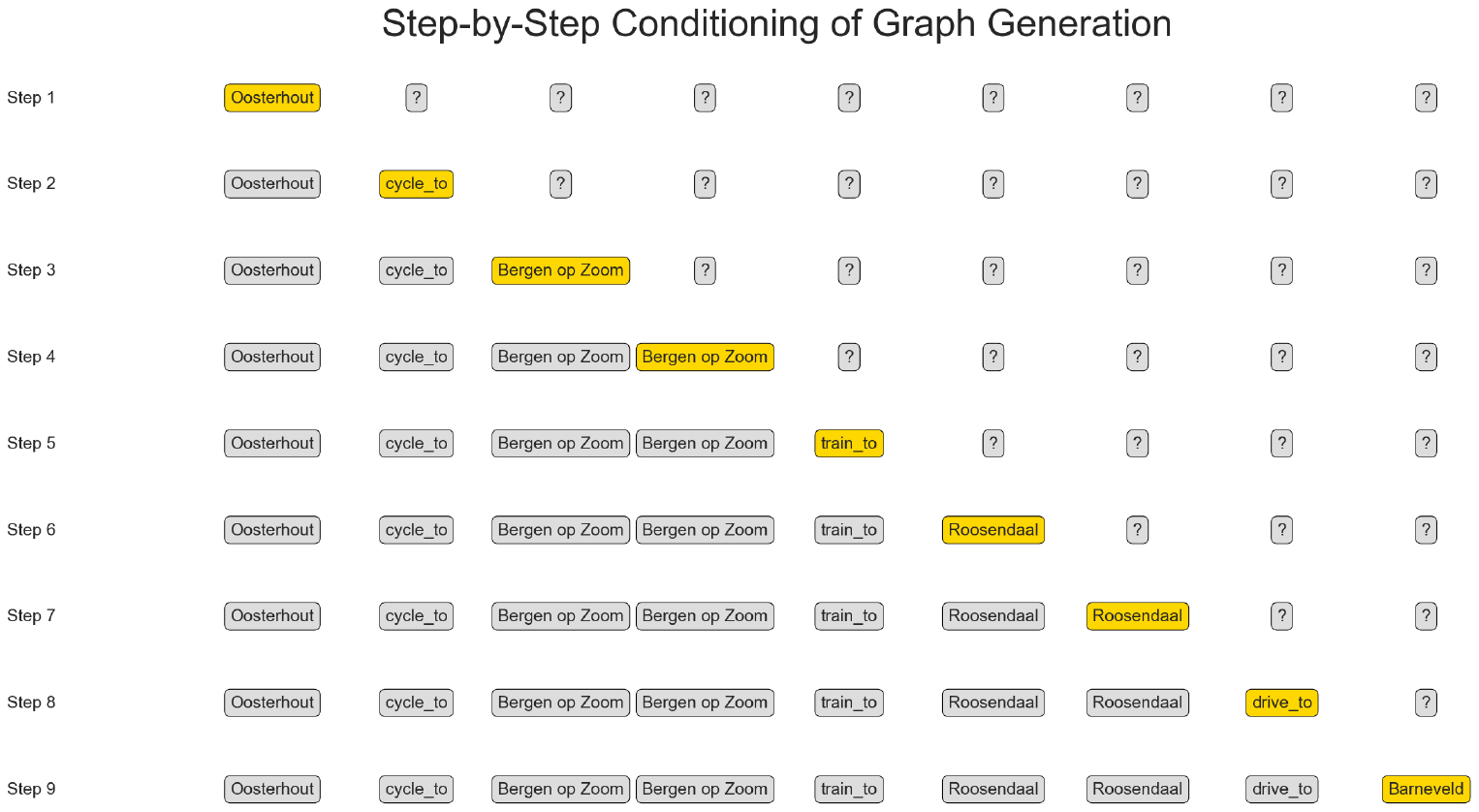}
        \caption{Example of syn-paths conditioned generation}
        \label{fig:conditioned_gen}
    \end{subfigure}
    
    \caption{Effect of progressive conditioning on Knowledge Graph generation for the syn-paths dataset. Subfigure (a) quantifies novelty and validity under increasing conditioning, (b) shows the corresponding reduction in sample diversity, and (c) provides an example of a conditioned generation where the model completes a partially specified graph.}
    \label{fig:cond-visualization}
\end{figure}

Figure \ref{fig:cond-visualization} shows that conditioned generation is also possible for the \modelname model, which allows the model to generate KGs and simultaneously enforces specific constraints. Entities or relations are fixed in place in the positions of interest, and then we decode the remaining tokens with constrained sampling (temperature/top-k/top-p). Figure \ref{fig:novel_constrain} shows that the novelty and validity of the generated structures remain high for all steps of the conditioning process, an indication that the model can produce triples and, consequently, graphs that are semantically correct. At the same time, as seen in Figure \ref{fig:diversity_constrain}, the diversity of the generated graphs drops dynamically as more entities and relations are added. This makes sense as the population of probable samples narrows with each additional constraint and limits the generative freedom of the model.

\subsection{Additional Comments about \modelname \& \modelvae}

\noindent \textbf{Variable Graph Length} \quad It is desirable to learn latent graph structures of varying sizes. In natural language processing, language models utilize special tokens to indicate the end of a sequence. Following a similar approach, we model variable-length KGs by linearizing graphs into a sequence of tokens and introducing boundary tokens. We always introduce \texttt{BOS} as the initial token and terminate generation upon emitting \texttt{EOS}, while using \texttt{PAD} for mini-batching. This simple setting allows the decoder to learn \emph{when} to stop and \emph{how large} the generated graphs should be, ensuring that the length distribution is learned. During inference, beam search halts on \texttt{EOS}, leading to the production of graphs of different sizes without any post hoc trim. In order to avoid length bias, we randomize triple order during training. In the probabilistic variant (\modelvae), the latent $\mathbf{z}$ conditions the entire sequence and this yields consistent length control across all samples, while at the same time preserving variability.

\newpage
\section{Additional Tables}

\begin{table}[!h]
\caption{Semantic validity of the graphs generated. We sample graphs and check the novelty of the sampled graphs by comparing them against the training and validation sets. The best performing models for each dataset are \textbf{bolded}. Baseline results are from the IntelliGraphs paper \citep{thanapalasingam2023intelligraphs}.}
\small
\label{graph-sampling-results_appendix}
\begin{center}
\begin{tabular}{llcccc}
\thickhline
\multirow{2}{*}{\textbf{Datasets}} & \multirow{2}{*}{\textbf{Model}} & \textbf{\% Valid} & \textbf{\% Novel} & \textbf{\% Novel} & \textbf{\% Empty} \\
& & \textbf{Graphs} $\uparrow$ & \textbf{\& Valid} $\uparrow$ & \textbf{Graphs} $\uparrow$ & \textbf{Graphs} $\downarrow$ \\
\thickhline
\multirow{7}{*}{\textbf{syn-paths}}
& uniform & 0 & 0 & 100.00 & 0 \\
& TransE & 0.25 & 0.25 & 23.45 & 76.55 \\
& DistMult & 0.69 & 0.69 & 14.59 & 85.41 \\
& ComplEx & 0.71 & 0.71 & 14.27 & 85.73 \\
& \transformerbaselineabbr & 99.60 & 99.60 & 100.00 & 0 \\
& \modelvae & 92.50 & 92.50 & 100.00 & 0 \\
& \transformerark & 97.39 & 97.39 & 100.00 & 0 \\
& \modelname  &  \textbf{99.95}& \textbf{99.95}& 100.00 & 0 \\
\hline
\multirow{7}{*}{\textbf{syn-tipr}}
& uniform & 0 & 0 & 100.00 & 0 \\
& TransE & 0 & 0 & 5.58 & 94.42 \\
& DistMult & 0 & 0 & 13.34 & 86.66 \\
& ComplEx & 0 & 0 & 4.95 & 96.05 \\
& \transformerbaselineabbr & 100.00 & 100.00 & 100.00 & 0 \\
& \modelvae & 98.45 & 98.45& 100.00 & 0 \\
& \transformerark & 100.00 & 100.00 & 100.00 & 0 \\
& \modelname  & \textbf{100.00}  &\textbf{100.00} & 100.00 & 0 \\
\hline
\multirow{7}{*}{\textbf{syn-types}}
& uniform & 0 & 0 & 100.00 & 0 \\
& TransE & 0.21 & 0.21 & 15.44 & 84.56 \\
& DistMult & 0.13 & 0.13 & 12.46 & 87.53 \\
& ComplEx & 0.07 & 0.07 & 10.25 & 89.75 \\
& \transformerbaselineabbr & 100.00 & \textbf{100.00} & \textbf{100.00} & 0 \\
& \modelvae & 100.00 & 100.00 & 100.00 & 0 \\
& \transformerark & 87.07 & 87.07 & 100.00 & 0 \\
& \modelname  & 89.22 & 89.22& 100.00 & 0 \\ 
\hline
\multirow{7}{*}{\textbf{wd-movies}}
& uniform & 0 & 0 & 100.00 & 0 \\
& TransE & 0 & 0 & 14.61 & 85.39 \\
& DistMult & 0 & 0 & 12.93 & 87.07 \\
& ComplEx & 0 & 0 & 1.87 & 98.13 \\
& \transformerbaselineabbr & \textbf{99.83} & \textbf{99.9} & \textbf{100} & 0 \\
& \modelvae & 99.47 &99.47 & 100.00 & 0 \\
& \transformerark & 98.33 & 98.33 & 100.00 & 0\\
& \modelname  & 99.24 & 99.24 & 100.00 & 0 \\
\hline
\multirow{7}{*}{\textbf{wd-articles}}
& uniform & 0 & 0 & 100.00 & 0 \\
& TransE & 0 & 0 & 4.58 & 95.42 \\
& DistMult & 0 & 0 & 0 & 100.00 \\
& ComplEx & 0 & 0 & 2.46 & 97.54 \\
& \transformerbaselineabbr & 98.00 & 98.00  & 100.00  & 0 \\
& \modelvae & \textbf{99.13} & \textbf{99.13} & 100.00 & 0 \\
& \transformerark & 95.37 & 95.37 & 100.00 & 0 \\
& \modelname  &97.24  &97.24 & 99.99 & 0 \\
\thickhline
\end{tabular}
\end{center}
\end{table}

\begin{table}[!h]
\caption{
    We measure the compression quality for compressing graphs $G$. $D_{KL}$ is only available for the VAE because it relies on the variational approximation, which is unique to this model. For the VAE, we compute an upper bound on the compression length (in bits). Probabilistic baseline (uniform, TransE, ComplEx, DistMult) results are from \citet{thanapalasingam2023intelligraphs}.
}
\small
\label{compression-results}
\begin{center}
\begin{tabular}{llcccc}
\thickhline
\multirow{2}{*}{\textbf{Datasets}} & \multirow{2}{*}{\textbf{Models}} & \multicolumn{4}{c}{\textbf{Compression Length (bits)}} \\
 & & \textbf{$G$} & \textbf{$S$} & \textbf{$E$} & \textbf{$D_{KL}$} \\
\thickhline
\multirow{7}{*}{\textbf{syn-paths}} 
 & uniform & 30.49 & 12.80 & 17.69 & - \\
 & TransE & 49.89 & 16.19 & 33.69 & - \\
 & ComplEx & 54.39 & 20.71 & 33.69 & - \\
 & DistMult & 48.58 & 14.90 & 33.69 & - \\
 & \transformerbaselineabbr & 27.77  & - & 14.47 & 13.30  \\
 & \modelvae & 28.74 & - & 18.41 &10.33  \\
  & \transformerark  & \textbf{27.57}  & -&- & - \\
 & \modelname  & 27.65 & -&- & - \\
\hline
\multirow{7}{*}{\textbf{syn-tipr}} 
 & uniform & 61.61 & 29.14 & 32.47 & - \\
 & TransE & 69.51 & 28.70 & 40.81 & - \\
 & ComplEx & 63.96 & 23.15 & 40.81 & - \\
 & DistMult & 67.51 & 26.70 & 40.81 & - \\
 & \transformerbaselineabbr & 26.30 & - & 11.13 & 15.17 \\
 & \modelvae & 27.14  & - & 9.90 & 17.24\\
 & \transformerark  & \textbf{23.34}& -&- & - \\
 & \modelname  & 23.48 &- &-  &-\\
\hline
\multirow{7}{*}{\textbf{syn-types}} 
 & uniform & \textbf{36.02} & 16.84 & 19.18 & - \\
 & TransE & 48.26 & 19.05 & 29.21 & - \\
 & ComplEx & 47.69 & 18.48 & 29.21 & - \\
 & DistMult & 47.46 & 18.24 & 29.21 & - \\
 & \transformerbaselineabbr & 59.61 & - & 59.46 & 0.15 \\
 & \modelvae & 60.58 & - & 60.37 & 0.21\\
 & \transformerark  & 59.79 & -&- & - \\
 & \modelname  & 59.63  &- &-  &-\\
\hline
\multirow{7}{*}{\textbf{wd-movies}} 
 & uniform & 171.60 & 53.86 & 117.74 & - \\
 & TransE & 208.60 & 51.39 & 157.21 & - \\
 & ComplEx & 202.68 & 45.46 & 157.21 & - \\
 & DistMult & 208.50 & 51.29 & 157.21 & - \\
 & \transformerbaselineabbr & 124.50 & - & 92.66 & 31.84 \\
 & \modelvae & 116.84 & - & 100.10 & 16.74 \\
 & \transformerark  & 114.49 & -&- & - \\
 & \modelname  & \textbf{98.19} & -& - &-\\
\hline
\multirow{7}{*}{\textbf{wd-articles}} 
 & uniform & 693.80 & 295.60 & 398.20 & - \\
 & TransE & 910.65 & 280.67 & 629.98 & - \\
 & ComplEx & 887.30 & 257.33 & 629.98 & - \\
 & DistMult & 901.91 & 271.94 & 629.98 & - \\ 
 & \transformerbaselineabbr &235.24  & - & 225.60 & 9.64 \\
 & \modelvae & \textbf{199.55}  & -&186.38  &13.17\\
 & \transformerark  & 224.25 & -&- & - \\
 & \modelname  & 205.24  & -& - &-\\
\thickhline
\end{tabular}
\end{center}
\end{table}

\begin{table}[!h]
\caption{
Latent space smoothness metrics for \transformerbaselineabbr\ and \modelvae\ with $\epsilon=0.1$. 
Higher local/global smoothness indicates more continuous transitions. Lower flip rates suggest larger regions mapping to identical graphs.
}
\label{tab:latent_smoothness_results_models}
\begin{center}
\begin{tabular}{llcccc}
\thickhline
\multirow{2}{*}{\textbf{Dataset}} & \multirow{2}{*}{\textbf{Model}} & \textbf{Local} & \textbf{Global} & \textbf{Flip} & \textbf{Avg Basin} \\
 & & \textbf{Smoothness} $\uparrow$ & \textbf{Consistency} $\uparrow$ & \textbf{Rate} $\downarrow$ & \textbf{Length} $\uparrow$ \\
\thickhline
\multirow{2}{*}{\textbf{syn-paths}}
 & \transformerbaselineabbr & 0.75 & 0.36  & 0.20 & 4.54 \\
 & \modelvae & 0.74 & 0.14 & 0.33 & 2.87 \\
\hline
\multirow{2}{*}{\textbf{syn-tipr}}
 & \transformerbaselineabbr & 0.99 & 0.98 & 0.09 & 8.61 \\
 & \modelvae & 0.93 & 0.69 & 0.10 & 8.03 \\
\hline
\multirow{2}{*}{\textbf{syn-types}}
 & \transformerbaselineabbr & 0.82 & 0.60 & 0.12 & 6.80 \\
 & \modelvae & 0.92 & 0.73 & 0.20 & 4.47 \\
\hline
\multirow{2}{*}{\textbf{wd-movies}}
 & \transformerbaselineabbr & 0.87 & 0.58 & 0.15 & 5.70 \\
 & \modelvae & 0.84 & 0.49 & 0.40 & 2.93 \\
\hline
\multirow{2}{*}{\textbf{wd-articles}}
 & \transformerbaselineabbr & 0.81 & 0.55 & 0.14 & 5.37 \\
 & \modelvae & 0.82 & 0.57 & 0.17 & 3.46 \\
\thickhline
\end{tabular}
\end{center}
\end{table}

\newpage
\newpage
\section{Additional Figures}

\subsection{Architectural Details}
\label{sec:architectural-details}

\begin{figure}[h]
    \centering
    \includegraphics[width=14cm]{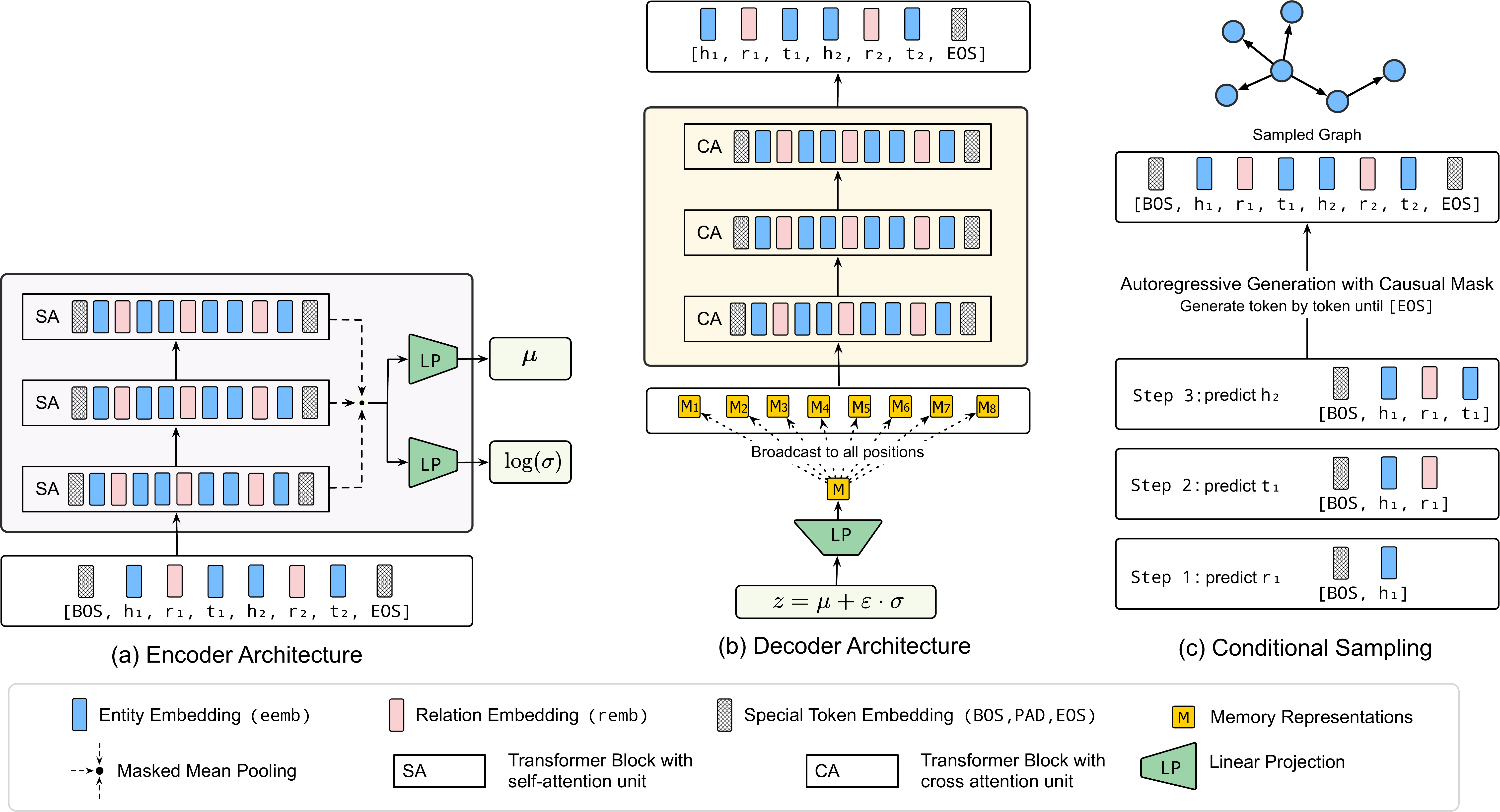}
    \caption{
    \transformerbaselineabbr has three main components: 
    (a) an \emph{Encoder} that processes linearized Knowledge Graph triple sequences $[\texttt{BOS}, h_1, r_1, t_1, h_2, r_2, t_2, \ldots, \texttt{EOS}]$ through self-attention (SA) blocks to produce latent distribution parameters $(\mu, \log\sigma)$, 
    (b) a \emph{Decoder} that uses cross-attention (CA) to condition on the sampled latent code $z$ and autoregressively generates token sequences with causal masking, and 
    (c) \emph{Conditional Sampling} that demonstrates the step-by-step autoregressive generation process, predicting one token at a time until the $\texttt{[EOS]}$ token is produced or the maximum sequence length is reached. The model uses a unified vocabulary embedding matrix spanning special tokens (\texttt{[BOS], [PAD], [EOS]}), entities (shown in blue), and relations (shown in pink), enabling sequential generation of Knowledge Graphs from learned latent representations.
    }
    \label{fig:transformer-baseline-architecture}
\end{figure}

Figure \ref{fig:transformer-baseline-architecture} shows the architectural details of the \transformerbaselineabbr model. Also Figure \ref{fig:director_generation_comparison} shows an example for conditioned generation.

\end{document}